\documentclass{article}

\usepackage{microtype}
\usepackage{graphicx}
\usepackage{subfigure}
\usepackage{booktabs} 
\usepackage{algpseudocode}

\usepackage{enumitem}
\setlist{topsep=0pt, leftmargin=*}
\usepackage{wrapfig}
 
\usepackage{tikz}
\usetikzlibrary{automata,arrows,positioning,calc}
\usetikzlibrary{arrows.meta, automata,
                positioning,
                quotes}

\usepackage{hyperref}

\usepackage[accepted]{icml2024}

\usepackage{amsmath}
\usepackage{amssymb}
\usepackage{mathtools}
\usepackage{amsthm}
\usepackage{dsfont}
\usepackage{caption}

\usepackage[capitalize,noabbrev]{cleveref}

\theoremstyle{plain}
\newtheorem{theorem}{Theorem}[section]
\newtheorem{proposition}[theorem]{Proposition}

\newtheorem{corollary}[theorem]{Corollary}
\theoremstyle{definition}
\newtheorem{definition}[theorem]{Definition}

\newtheorem{remark}{Remark}

\theoremstyle{remark}

\usepackage[textsize=tiny]{todonotes}
\usepackage{amssymb}
\usepackage{nccmath}
\AtBeginEnvironment{align*}{\useshortskip}

\definecolor{myyellow}{HTML}{FEE483}
\definecolor{mylyellow}{HTML}{FFF5CC}
\definecolor{mygreen}{HTML}{A0E38D}
\definecolor{mylgreen}{HTML}{D8EB9F}
\definecolor{myblue}{HTML}{1971C2}
\definecolor{myred}{HTML}{E32636}

\icmltitlerunning{Understanding and Mitigating Tokenization Bias in Language Models}

\begin{document}

\twocolumn[
\icmltitle{Understanding and Mitigating Tokenization Bias in Language Models}


\icmlsetsymbol{equal}{*}

\begin{icmlauthorlist}
\icmlauthor{Buu Phan}{equal,yyy}
\icmlauthor{Marton Havasi}{comp}
\icmlauthor{Matthew Muckley}{comp}
\icmlauthor{Karen Ullrich}{comp}
\end{icmlauthorlist}

\icmlaffiliation{yyy}{University of Toronto}
\icmlaffiliation{comp}{Meta AI}

\icmlcorrespondingauthor{Buu Phan}{truong.phan@mail.utoronto.ca}
\icmlcorrespondingauthor{Karen Ullrich}{karenu@meta.com}

\icmlkeywords{Machine Learning, ICML}

\vskip 0.3in
]



\printAffiliationsAndNotice{\icmlEqualContribution} 

\begin{abstract}
State-of-the-art language models are autoregressive and operate on subword units known as tokens. Specifically, one must encode the conditioning string into a list of tokens before passing to the language models for next-token prediction. We show that popular encoding schemes, such as maximum prefix encoding (MPE) and byte-pair-encoding (BPE), induce a sampling bias that cannot be mitigated with more training or data. To counter this universal problem, for each encoding scheme above, we propose a novel algorithm to obtain unbiased estimates from any language model trained on tokenized data. Our methods do not require finetuning the model, and the complexity, defined as the number of model runs, scales linearly with the sequence length in the case of MPE. As a result, we show that one can simulate token-free behavior from a tokenized language model. We empirically verify the correctness of our method through a Markov-chain setup, where it accurately recovers the transition probabilities, as opposed to the conventional method of directly prompting tokens into the language model. 
\end{abstract}

\section{Introduction}
\label{intro}

Tokenization is a preprocessing procedure used in many state-of-the-art (SOTA)  language models (LMs) such as GPTs \cite{brown2020language}, Llama \cite{touvron2023llama} and Gemini \cite{team2023gemini}.
It divides the input text into smaller subword units while retaining linguistic importance, helping to address vocabulary limitations such as unknown words.
Tokenization also shortens (compresses) the input context length \cite{sennrich2015neural,kudo2018sentencepiece}.
Since effective compression allows transformer-based LMs to handle longer context strings, many works \cite{zouhar2023tokenization,galle2019investigating,goldman2024unpacking} have focused on enhancing vocabulary design and encoding algorithms for better performance in downstream tasks. However, the relationship between compression and model performance remains unclear.
Some research suggests the impact of compression is not always positive \cite{schmidt2024tokenization, dagan2024getting, goyal2023think}. Consequently, understanding tokenization's effect on model performance continues to be an open question.

Tokenization has been criticized for introducing many shortcomings in LMs. These include sensitivity to spelling and morphological structure \cite{xue2022byt5}, language-based biases \cite{petrov2024language}, subpar performance in specific tasks such as arithmetic \cite{singh2024tokenization}, or new domains \cite{liu2023task}. One approach to address these issues is through fine-tuning the model with new vocabularies;  however, this often complicates the training process and requires domain-specific expertise \cite{chen2023improving,liu2023ofa}. Furthermore, the performance gains do not provide a theoretical understanding of whether these limitations truly arise from the tokenization process or result from suboptimal model training. Another direction is to develop token-free LMs \cite{yu2024megabyte,nawrot2022efficient,tay2021charformer}. While this approach has potential as it eliminates tokenization-related issues, it significantly increases the context length, resulting in performance that still lags behind the SOTA tokenized LMs \footnote{We refer language models that process tokenized texts as tokenized language models (tokenized LMs).}\cite{yu2024megabyte}.

In this work we offer new theoretical insights on the behavior of tokenized LMs.
We show that they are statistically equivalent to their token-free counterparts. Specifically, we examine the maximum prefix encoding (MPE) scheme employed in the WordPiece tokenization method \cite{devlin2018bert, song2020fast} and find that this process not only  results in biased estimates of next token probabilities, but also leads to overall skewed estimates of subsequent character probabilities. In general, this bias persists despite an increase in training data, even within the simple setting of a 1st-order Markov chain. Such bias occurs due to the implicit disparity between the domain of the conditioning context, namely, characters versus tokens. Nevertheless, we will show that it is possible to correct this bias without resorting to finetuning. Once adjusted, it becomes possible to simulate the token-free behavior learned implicitly by the tokenized LM and even (theoretically) mimic the behavior of another tokenized model employing a distinct vocabulary set, all without requiring finetuning. Our specific contributions are as follows: \begin{itemize}\itemsep0em 
    \item We show the presence of a bias in the next-token distribution that arises as a result of the tokenization process. 
    \item We present two novel algorithms to correct this bias for MPE and Byte-Pair-Encoding (BPE) respectively. Due to space limit, the analysis and algorithm for BPE are presented in Appendix \ref{bpe}. 
    \item We verify the correctness of our algorithms on learning the transition matrix of a $k$-th order Markov chain.
\end{itemize}



\vspace{-9pt}
\section{Problem Setup}\label{bias_analysis} \vspace{-5pt}


\begin{figure*}[t]
    \centering
    \vspace{-12pt}
    \begin{tikzpicture}[->, >=stealth', auto, semithick, node distance=1.5cm]
	\tikzstyle{every state}=[fill=white,draw=black,thick,text=black,scale=1]
	\node[state, fill=green,  opacity=.0, draw=black,thick,  text opacity=1, minimum size=0.5cm,inner sep=0pt, outer sep=0pt]  (X)  {\footnotesize};
 \node[state, fill=green,  opacity=.3, draw=black,thick,  text opacity=1, minimum size=0.5cm,inner sep=0pt, outer sep=0pt]  (A)[below = 0.4cm of X]  {\footnotesize$A$};
	\node[state, fill=orange, opacity=.5, draw=black,thick,  text opacity=1, minimum size=0.5cm,inner sep=0pt, outer sep=0pt]    (B)[right of=A]   {\footnotesize$B$};
 \node[state, fill=green,  opacity=.0, draw=black,thick,  text opacity=1, minimum size=0.5cm,inner sep=0pt, outer sep=0pt]  (C)[below = 1.2cm of X]  {\footnotesize};
 \node[][below right = 0.1cm and -0.8cm of C] (M1) {\footnotesize \textcolor{white}{Before Tokenization }}; 
 \node[][below = -0.05cm of M1] (M2) {\footnotesize \textit{Input String:} $``AABABAAAABAAB"$ $\textcolor{white}{|}$};
	\path
	(A) edge[loop left]			node{\scriptsize$1 {-} \alpha$}	(A)
	(B) edge[bend left,below]	node{\scriptsize$\beta$}	(A)
        (A) edge[bend left,above]   node{\scriptsize$\alpha$}	(B)
        (B) edge[loop right]		node{\scriptsize$1 {-} \beta$}	(B);
	\end{tikzpicture}\hspace{-0.15cm} 
 \begin{tikzpicture}
     \node [shape=rectangle, align=center](table2) {
            \small
            \begin{tabular}{|c|c|}
            \hline
                ID & Token  \\
                \hline 
                1 &$ A$ \\
                \hline 
                2 & $B$ \\
                \hline
                3 & $AA$ \\
                \hline
            \end{tabular}
        };
        \node[][above right = -0.1cm and -2.45cm of table2] {\footnotesize Token Vocabulary};
        \node[shape=rectangle,draw=black][below = 0.25cm of table2] (bpe) {\footnotesize WordPiece Encoding };
        \node[][left = 0.0cm of bpe] {\footnotesize $\longrightarrow$};
        \node[][right = 0.0cm of bpe] {\footnotesize $\longrightarrow$};
 \end{tikzpicture}\hspace{-0.1cm}
	\begin{tikzpicture}[->, >=stealth', auto, semithick]
	\tikzstyle{every state}=[fill=white,draw=black,thick,text=black,scale=1]
	\node[state, fill=blue,  opacity=.3, draw=black,thick,  text opacity=1, minimum size=0.5cm,inner sep=0pt, outer sep=0pt]  (C)  {\footnotesize$AA$};
 \node[state, fill=green, opacity=.3, draw=black,thick,  text opacity=1, minimum size=0.5cm,inner sep=0pt, outer sep=0pt]    (A)[below right = 1.2cm and 0.6cm of C]   {\footnotesize$A$};
	\node[state, fill=orange, opacity=.5, draw=black,thick,  text opacity=1, minimum size=0.5cm,inner sep=0pt, outer sep=0pt]    (B)[right = 1.25cm of C]   {\footnotesize$B$};
 \node[][below = -0.05cm of M1] (M2) {\footnotesize \textit{Output Tokens:} $``\textcolor{blue}{AA}|\textcolor{orange}{B}|\textcolor{green}{A}|\textcolor{orange}{B}|\textcolor{blue}{AA}|\textcolor{blue}{AA}|\textcolor{orange}{B}|\textcolor{blue}{AA}|\textcolor{orange}{B}"$};
	\path
	(B) edge[,left]	node{\scriptsize$\alpha\beta$}	(A)
        (A) edge[red][bend right,right]   node{\scriptsize$1.0$}	(B)
        (A) edge[pink][left, right]   node[pos=0.65]{\scriptsize$0.0$}	(C)
        (A) edge[pink][out=240,in=300, loop, right,right] node[pos=0.8]{\scriptsize$0.0$} (A)
        (C) edge[out=170,in=120, loop, left] node{\scriptsize${(}1{-}\alpha{)}^{2}$} (C)
        (B) edge[,below]   node{\scriptsize$(1{-}\alpha)\beta$} (C)
        (C) edge[bend right,left]   node{\scriptsize$\alpha{(}1{-}\alpha{)}$} (A)
        (C) edge[bend left,above]   node{\scriptsize$\alpha$} (B)
        (B) edge[in=10,out=60, loop, right] node{\scriptsize$1{-}\beta$}	(B);
	\end{tikzpicture}
 \vspace{-5pt}
 \captionsetup{font=footnotesize}

 \caption{Next-Character sampling bias introduced by the WordPiece encoding algorithm. In this example, given the context token $``A"$, the model will always predict the next token as $``B"$ with probability $1.0$. We present a technique that, given a language model trained on tokenized domain, eliminate this bias and recover the accurate unbiased sampling distribution.}
 \label{markovchange}
 \vspace{-14pt}
\end{figure*}
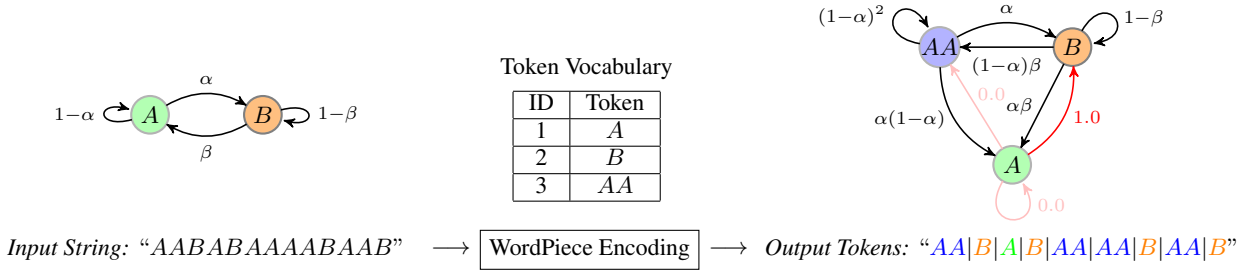
We begin by establishing the tokenization and language models setup in our paper. We then describe the next-character sampling bias problem due to tokenization. 


\vspace{-8pt}
\subsection{Notations and Setup. }\label{tok_setup}
 \textbf{String Notations. }For any string $s$, we denote its substring from $i$ to $j$ as $x^j_i{\vcentcolon=}{x_ix_{i+1}..x_j}$,  where each $x$ is a character of the alphabet $\mathcal{A}$. For a given string $x^N_1$, we define the prefix function that generates a set containing all possible prefix strings of $x^N_1$, represented as $\mathrm{prefix}(x^N_1) {=} \{x^1_1, x^2_1, x^3_1, ..., x^N_1\}$. Also, we define a concatenation function $\mathrm{concat}(.)$  that concatenates the given  list of  strings, e.g given $s_1{=}x^{N_1}_1$ and $s_2{=}y^{N_2}_1$, we obtain  $\mathrm{concat}(s_1,s_2) {=} \mathrm{concat}(x^{N_1}_1,y^{N_2}_1){=}x_1...x_{N_1}y_1...y_{N_2}$. Finally, we denote the set of all strings that start with a prefix $x^n_1$ as $\mathcal{S}(x^n_1){=}\{s|x^n_1{\in} \mathrm{prefix}(s)\}$.

\textbf{Tokenization Setting. } We assume having a predefined vocabulary $\mathcal{V}$ constructed using any tokenization algorithm such as BPE, with the condition that $\mathcal{A} {\subseteq }\mathcal{V}$.  We use $t$ to denote a token in $\mathcal{V}$, i.e. $t{\in} \mathcal{V}$. Importantly, we use the longest prefix matching strategy for tokenization (encoding), denoted as $\mathrm{encode}(.)$, similar to the approach used in the Wordpiece algorithm \cite{devlin2018bert, song2020fast}. Given a sequence of tokens $t^k_1$, the function $\mathrm{decode}(.)$ returns the concatenated string resulting from processing each token in the sequence. Finally, {the set of all strings that starts with the tokens $t^k_1$ is defined as $\mathcal{S}(t^k_1){=}\{s|t^k_1 {=} \mathrm{encode}(s)^k_1\}$.} 

\textbf{Tokenized LMs. } We assume having access to a tokenized autoregressive LM with parameters $\theta$ that is trained with tokens from $\mathcal{V}$ and maximum prefix matching. The target distributions on the character domain is denoted as $P_{\mathrm{gt}}(x^N_{n+1}|x^n_1)$ and on the token domain is $P_{\mathrm{gt}}(t_{i+1}|t^i_1)$. For simplicity, unless otherwise stated, we implicitly assume each probability term involves $\theta$. Using the model, we assume that one can compute $P(t_{i+1}|t^i_1)$ for any integer $i>0$. In this work, we consider LMs trained under the standard setup, where each string $s$ in the dataset is first tokenized with the encoding function $\mathrm{encode}(.)$ and vocabulary $\mathcal{V}$, and the parameters $\theta$ are optimized to maximize the predictive likelihood of the next token in the tokenized dataset. 
\vspace{-10pt}
\subsection{Next-Character Sampling Bias}\vspace{-3pt}
We first define the (next-character) sampling bias problem that describes the discrepancy between the character level and token level predictions for tokenized LMs. 
\begin{definition}\label{def_sampbias}
(Next-Character Sampling Bias) Let the input prompt string $x^{n}_1$ has $t^i_1{=}\mathrm{encode}(x^{n}_1)$ as the corresponding encoding. The next-character sampling bias occurs for this prompt when $P_{\mathrm{gt}}(x_{n+1}|x^{n}_1){\neq}P_{\mathrm{gt}}(x_{n+1}|t^{i}_1)$ where  $P_{\mathrm{gt}}(x_{n+1}|t^{i}_1){=}\sum_{t\in\mathcal{E}} P_{\mathrm{gt}}(t_{i+1}{=}t|t^i_1)$ where $\mathcal{E}=\{ t \in \mathcal{V} | \mathrm{decode}(t)\in\mathcal{S}(x_{n+1})\}$. 
\vspace{-5pt}
\end{definition}In other words, the probability of the next character being
``c" may be different from the sum of the probabilities of all tokens that start with ``c". Note that this character-level probability offers a broader perspective compared to the probability of the subsequent token being exactly ``c".



\textbf{Example. }Consider  a first order Markov chain with two states $\{``A",``B"\}$ as shown in Figure \ref{markovchange} (left). Each string is tokenized with $\mathcal{V}{=}\{``AA",``A",``B"\}$, which leads to a new Markov chain whose states and transition matrix is shown in Figure \ref{markovchange} (right). Details on computing the transition matrix of the new Markov chain is in Appendix \ref{tkfree2vocab}. We first observe that for the prompt $s_1{=}``AA"$ and $s_2{=}"B"$, there is no bias problem after marginalization\footnote{For example, we have $P_\mathrm{gt}(t_{i{+}1}{=}``AA"|t_{i}{=}``AA")+P_\mathrm{gt}(t_{i{+}1}{=}``A"|t_{i}{=}``AA")=\alpha=P_\mathrm{gt}(x_{n{+}1}{=}``A"|x_{n}{=}``A")$}. However, for the prompt $s_3{=}``A"$, the sampling bias occurs as $P_\mathrm{gt}(x_2{=}``B"|t_{1}{=}``A"){=}1.0$, which is not equal to $P_\mathrm{gt}(x_{2}{=}``B"|x_{1}{=}``A"){=}\alpha$, i.e. the optimally trained LM will always output $``B"$. In fact, for any context string that ends with token $``A"$, e.g $``AA|A"$ and $``B|A"$ (tokens are separated by $``|"$), such LM will always output $``B"$. 


Since this applies to any optimally trained LM, increasing the training set size does not mitigate this problem. The reason for this sampling bias is that, during the tokenization process with longest prefix matching, the token $``A"$ must be followed by the token $``B"$. Else, MPE encoding will merge to create a longer token $``AA"$. We generalize this phenomenon with the definition of {invalid encodings.} 



\begin{definition}
    (Invalid Encodings) The list of tokens (an encoding) $t^i_1$ is invalid if $\mathrm{encode}(\mathrm{decode}(t^i_1)) {\neq} t^i_1$. Otherwise, it is a valid encoding. 
\end{definition}
For example, let $\mathcal{V}{=}\{``c",``a",``t",``at",``cat"\}$ then $[``c"{,}``at"{,}``t"]$ and $[``c"{,}``a"{,}``t"{,}``t"]$ are invalid encodings of $``catt"$.
 We now show in Proposition \ref{undef_prob_main} that the existence of invalid encodings introduces sampling bias, generalizing the observed phenomenon in the Markov chain example to any autoregressive distribution.

\begin{proposition}\label{undef_prob_main} (Token-Induced Zero Probability)
    Let $t^i_1$ be a sequence of input tokens. For any invalid encoding $t^i_1$, we have $P_{\mathrm{gt}}(t^i_1){=}0.0$ and the conditional probability $P_{\mathrm{gt}}(t_{i+1}|t^i_1)$ is undefined. In the case $t^i_1$ is valid, then $P_{\mathrm{gt}}(t_{i+1}|t^i_1){=}0.0$ if $t^{i+1}_1$ is invalid.  Furthermore, let $x^{n}_1{=}\mathrm{decode}(t^i_1)$, then for any string $x^{N}_{n+1}$ such that $\mathrm{encode}(\mathrm{concat}(\mathrm{decode}(t^i_1), x^{N}_{n+1})) {\neq} t^i_1$, we have $P_{\mathrm{gt}}(x^{N}_{n+1}|t^i_1){=}0.0.$
\end{proposition}\begin{proof}\vspace{-5pt}
    See Appendix \ref{undef_prob_main_proof}.
    \vspace{-3pt}
\end{proof}

\begin{remark}
    Proposition \ref{undef_prob_main} implies that LMs may not function as expected when presented with invalid encodings, because these models will never be exposed to such inputs within the dataset. This directly implies that the practice of evaluating LMs under different encodings  \cite{cao2021you,chirkova2023should} is suboptimal.

    
\end{remark}

\vspace{-15pt}
\section{Alleviating Sampling Bias} \label{method}
\vspace{-5pt}
We propose a method to remove the described bias and recover the original token-free autoregressive model, i.e. expressing the implicitly learned $P(x^N_{n+1}|x^n_1)$ using the tokenized LM that outputs the conditional probability $P(t_{i+1}|t^i_1)$. For $N{=}n{+}1$, this captures the behavior of a token-free model, i.e. sampling the next character instead of a whole token. {We assume our LM follows Proposition \ref{undef_prob_main} on zero probability events and undefined conditional probability for invalid encodings. Appendix \ref{lm_justify} justifies this assumption and provides its practical implementation.} 

Our method consists of two stages. In the first stage, the idea is to identify the condition when $P(x^N_{n+1}|t^i_1) = P(x^N_{n+1}|x^{n}_1)$ where $t^{i}_1=\mathrm{encode}(x^{n}_1)$. Once identified, we can refactor the conditional probability to match the conditioning events. In the second stage, we compute $P(x^N_{n+1}|t^i_1)$ using the LM output probability, i.e. $P(t_{i+1}|t^i_1)$, through the novel Maximum Prefix Correction (MPC) Algorithm.



\vspace{-8pt}
\subsection{Refactoring}
\vspace{-3pt}

Our method removes the bias by connecting character and token domains through a special subset of tokens $\mathcal{V}^* {\subset} \mathcal{V}$, whose elements $t^* {\in} \mathcal{V}^*$ are not a substring of any other tokens in $\mathcal{V}$ but itself. For example, given $\mathcal{V}{=}\{``AAA", ``AA", ``CB", ``A", ``B", ``C"\}$, then $\mathcal{V}^*=\{``AAA", ``CB"\}$. In the Markov example in Section \ref{bias_analysis}, this corresponds to the tokens $``AA"$ and $``B"$. Also, we assume that any string $x^N_1$ has the first token $t_1 \in \mathcal{V}^*$\footnote{Many current language models begins with a start token ${<}\mathrm{start}{>}$ in $\mathcal{V}^*$, e.g.   in SentencePiece \cite{kudo2018sentencepiece}.}. Consider the input string $x^{n}_1$ and its corresponding encoding $t^i_1{=}\mathrm{encode}(x^{n}_1)$, Proposition \ref{invert_main} shows the sufficient condition for $\mathcal{S}(t^i_1){=}\mathcal{S}(x^{n}_1)$.

\begin{proposition}\label{invert_main} 
    Let $s^*=x^{n}_1$, where $t^i_1=\mathrm{encode}(s^*)=\mathrm{encode}(x^{n}_1)$. Then we have $\mathcal{S}(t^i_1) \subset \mathcal{S}(x^{n}_1)$, i.e. for any string $s$ where $t^i_1=\mathrm{encode}(s)^i_1$,  we have $P(x^{n}_1|t^i_1)=1.0$. In the case $t_i \in \mathcal{V}^*$, then we also have that $\mathcal{S}(t^i_1) = \mathcal{S}(x^{n}_1)$, i.e. any string $s$ where $x^{n}_1\in \mathrm{prefix}(s)$ must have the first $i$ tokens as $t^i_1$ and $P(t^i_1|x^{n}_1)=1.0$. 
\end{proposition}
\begin{proof} \vspace{-8pt}
    See Appendix \ref{backup_main_proof}.
     \vspace{-8pt}
\end{proof}
The intuition for Proposition \ref{invert_main} is that the subsequent string after $t_i{\in}\mathcal{V}^*$ cannot change the tokenization for $x^n_1$. We now establish one of the main results in Corollary \ref{backup_equality_main}.

\begin{corollary}\label{backup_equality_main}
    Following Proposition \ref{invert_main}, suppose $t_i \in \mathcal{V}^*$ then we have $P(x^N_{n+1}|x^{n}_1{)}  {=} P(x^N_{n{+}1}|t^i_1)$. Similarly, we also have $P(t^j_{i+1}|x^{n}_1{)}  {=} P(t^j_{i+1}|t^i_1)$.
\end{corollary}
\begin{proof}\vspace{-8pt}
    See Appendix \ref{backup_main_proof}. 
     \vspace{-8pt}
\end{proof}
We note that Proposition \ref{invert_main} and Corollary \ref{backup_equality_main} always hold, regardless of the value of $\theta$. In general, consider when the last token of $\mathrm{encode}(x^n_1)$ is not in $\mathcal{V}^*$, we can refactor  $P(x^N_{n+1}|x^n_1)$ as follow:
\vspace{-8pt}\begin{align}
    P(x^N_{n+1}|x^n_1)  
    {=} \frac{P(x^N_{n_k+1}|t^k_1)}{P(x^n_{n_k+1}|t^k_1)},  \label{fractorial_main}
\end{align}
where $k$ is the last token in $\mathrm{encode}(x^n_1)$ such that $t_k {\in} \mathcal{V}^*$ and $x^{n_k}_1{=}\mathrm{decode}(t^k_1)$, where $n_k \leq n$. Proof details of this step can be found in the Appendix \ref{BP_proof}. We then use the MPC algorithm to compute each term in the RHS individually. 

\vspace{-10pt}
\subsection{Maximum Prefix Correction Algorithm}
\label{algorithm}

\setlength{\textfloatsep}{3pt}

\vspace{-5pt}\begin{algorithm}[t]\footnotesize

\captionsetup{font=footnotesize}
\caption{\footnotesize{Maximum Prefix Correction  Algorithm. This algorithm recursively computes $P(x^N_{n_k+1}|t^k_1)$.}}
\label{algo:BP}
\begin{algorithmic}[1]

\Procedure{compute}{$x^N_{n_k+1},t^k_1$}       
    \State //\textit{ Branching Step:}

\State $\mathcal{B}=\{t\in\mathcal{V}|x^N_{n_k {+} 1} {\in} \mathrm{prefix}( \mathrm{decode} ({t}))\}$
 \State $\mathrm{b_{val}}={\sum\limits_{t {\in }\mathcal{B}}} P(t_{k{+}1}=t \big{|} t^k_1)$ 
    
    \State //\textit{ Base Case:}
    \If{$ \mathrm{encode}(x^N_i) \in \mathcal{V}$} 
    \State \Return $\mathrm{b_{val}}$
    \EndIf
    \State //\textit{Extract the Next Token:}
    \State $t_{k+1} =\mathrm{encode}(x^N_{n_k+1})_1$
    \State //\textit{ Passing Step:}
    \State $\mathrm{p_{val}} =   P( t_{k+1} \big{|} t^k_1) $ 
    \State $\mathrm{p_{val}} =  \mathrm{p_{val}} \times \text{COMPUTE}(x^N_{n_{k+1}+1} ,t^{k+1}_1)  $
\State    \Return $\mathrm{b_{val}} + \mathrm{p_{val}}$ 
\EndProcedure
\end{algorithmic}

\end{algorithm}


We present the MPC algorithm  in Algorithm \ref{algo:BP}, that allows us to compute the probabilities  $P(x^N_{n_k+1}|t^k_1)$ and $P(x^n_{n_k+1}|t^k_1)$ in Equation (\ref{fractorial_main}). Note that this algorithm does not require $t_k{\in}\mathcal{V}^* $.  Details on the algorithmic correctness are shown in Appendix \ref{BP_proof}. 

The idea is to marginalize out $P(x^N_{n_k+1}|t^k_1)$ by considering two complementary events: when the next token $t_{k+1}$ has a prefix $x^N_{n_k+1}$ ($\mathrm{b_{val}}$ in the Branch Step) versus when the next token $t_{k+1}$ is contained within $x^N_{n_k+1}$ ($\mathrm{p_{val}}$ in the  Pass Step). Formally, MPC computes the  following probabilities:\begin{align}
\vspace*{-20pt}
    \mathrm{b_{val}} &= P(x^N_{n_k {+} 1},  t_{k+1} \in \mathcal{B}(x^N_{n_k+1})) \big{|}  t^k_1), \\
    \mathrm{p_{val}} &= P(x^N_{n_k {+} 1},  t_{k+1} \notin \mathcal{B}(x^N_{n_k+1})) \big{|}  t^k_1),
\vspace{-20pt}
\end{align} where $\mathcal{B}(x^N_{n_k+1}) {=} \{t{\in}\mathcal{V}|x^N_{n_k {+} 1} {\in} \mathrm{prefix}( \mathrm{decode} ({t}))\}$ and we immediately see that $P(x^N_{n_k+1}|t^k_1){=}\mathrm{b_{val} {+} p_{val}}$. 

\begin{figure}[t]
\vspace{-0pt}
\begin{center}
\centerline{\includegraphics[width=\columnwidth]{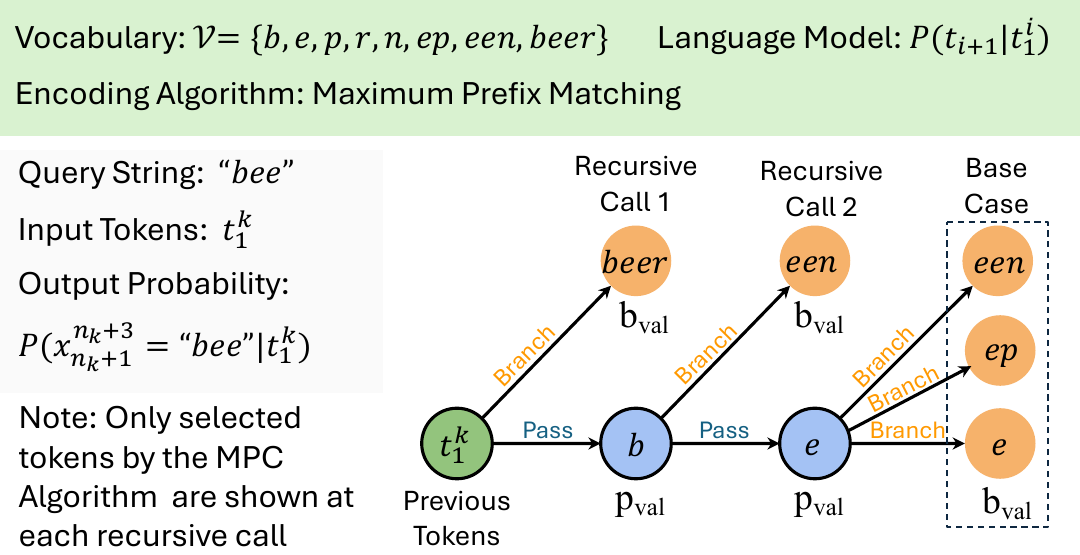}} 
\vspace{-2pt}
\captionsetup{font=footnotesize}
\caption{MPC Visualization. At each recursive call, the Branch step finds tokens that starts with the query string while the Pass step extracts and employs the next token and leftover string  for the next recursive call until meeting  the base case.}

\label{algo_visual}
\end{center}
\vspace{-10pt}
\end{figure}

We provide an intuitive explanation for the algorithm following the example in Figure \ref{algo_visual}. Here, we would like to compute the probability $P(x^{n_k+3}_{n_k+1}{=}``bee"|t^k_1)$. The first possibility is that $``bee"$ is a prefix of the next token, so we search for all such tokens (line 3 in the algorithm) and sum up their probability (line 4), i.e. $\mathrm{b_{val}}{=}P(t_{k+1}{=}``beer"|t^k_1)$. Figure \ref{algo_visual} visualizes this step as branching out the tree by finding all tokens completing the string. Since $``beer"$ is not the only string that contains $``bee"$, e.g. $``beep", ``been"$, etc. we need to compute the probability for these other scenarios, each of which has $t_{k+1}{=}``b"$ (the first token in $``bee"$, line 10 and 12) due to maximum prefix encoding. Then, we want to compute the probability that the subsequent string is $``ee"$ (line 13), given the previous $t^k_1$ and $t_{k+1}{=}``b"$, which is the output of the MPC algorithm but for $x^{n_k+3}_{n_k+2}{=}``ee"$ and $t^{k+1}_1$. Formally, in the Passing step: $\mathrm{p_{val}}{=}P(t_{k+1}{=}``b"|t^k_1)P(x^{n_k+3}_{n_k+2}{=}``ee"|t^k_1, t_{k+1}{=}``b")$. We continue the procedure until meeting the base case, where the string must be a prefix of the next token (usually, when there is only a single character left). Finally, by computing the sum of the branch and pass steps, we obtain the desired conditional probability $\mathrm{b_{val}}{+}\mathrm{p_{val}}{=}P(x^{n_k+3}_{n_k+1}{=}``bee"|t^k_1)$. 

\vspace{-5pt}
\section{Experiments}\label{mpe_exp}\vspace{-5pt}

\begin{figure}[t]
\vspace{-10pt}
\begin{center}
\centerline{\includegraphics[width=\columnwidth]{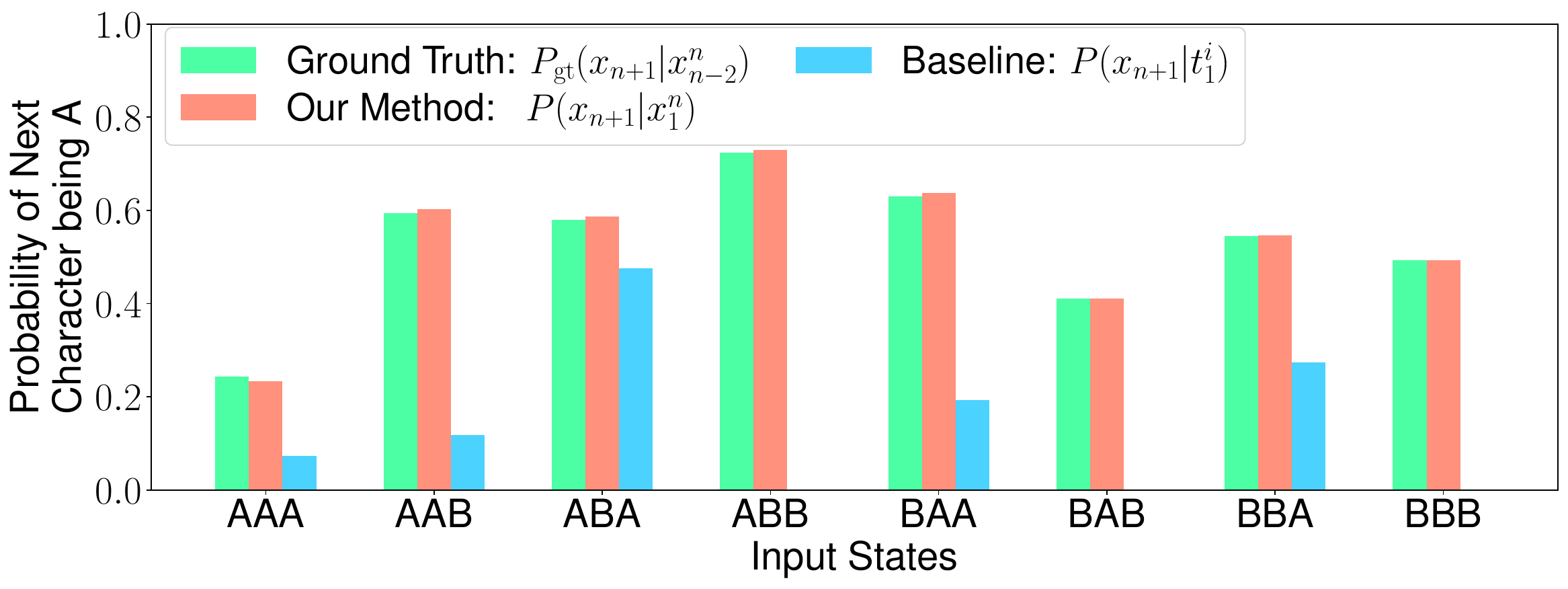}}
\vspace{-5pt}
\captionsetup{font=footnotesize}
\caption{Our method accurately estimates the transition probability of a 3rd order Markov chain while the baseline method fails to.}
\label{results}
\end{center}
\vspace{-12pt}
\end{figure}

We validate our method on a 3rd order Markov chain experiment  with $\mathcal{A}{=}\{``A"{,}``B"\}$, where we randomly construct  the transition matrix and the vocabulary
$\mathcal{V}{=}\{``A",``B",``AA",``BAAB",``BBAA",``BBBA" , $ $``BA", ``BBA"\}$.  We train a LM model using GPT-2 architecture with 6 hidden layers. Since the model is agnostic to the Markov chain order, we average the probability from 100 runs on different context length while fixing the last 3 characters.  We compare our method with the baseline estimator $P(x_{n+1}|t^i_1)$, equivalent to one Branch step in the MPC algorithm. Figure \ref{results} shows the results where the baseline method exhibits significant sampling bias due to tokenization. {Following Proposition \ref{undef_prob_main}, one can clarify the zero probability events output from the baseline estimator.} Our method, in contrast, accurately estimates the ground truth probability used to generate the data, showing that it is possible to recover the implicitly learned character information from the tokenized LMs. 
\vspace{-10pt}

\section{Conclusion}\vspace{-5pt}
This work identifies the next-character sampling gap between a tokenized model and a token-free one, which persists even for optimally trained models. We present a probabilistic approach to effectively eliminate this bias without requiring additional training. This closes the sampling gap between tokenized and token-free models, suggesting that language models implicitly absorb character-level information despite being trained solely on tokenized text. This result implies that it is theoretically possible to simulate the behavior of another language model trained using different vocabulary without any fine-tuning, since it is possible to transfer from token-free models to tokenized counterparts. 


\newpage
\bibliography{example_paper}
\bibliographystyle{icml2024}

\newpage
\appendix
\onecolumn
\section{Related Work}
\label{related_work}

\textbf{Theory of Tokenization. }Existing works on tokenization  generally support the idea that compressing tokens enhances model performance \cite{galle2019investigating,gutierrez2023languages,zouhar2023tokenization}. However, these emperically findings  are in conflicted with other later studies \citet{cognetta2024two,schmidt2024tokenization}. On the theoretical side, \citet{rajaraman2024toward} examined tokenization through the lens of unigram models, motivated by the observation made by \citet{makkuva2024attention} that transformers struggles to learn 2nd-order Markov chains.  We, however, do not observe this phenomenon in our experiment. As such, our work on bias due to tokenization is not affected by their observation. 

\textbf{Tokenization and Perplexity.} Our work relates to the statistical evaluation of LMs, where we provide an algorithm to directly evaluate the character-level perplexity $p(x^{N}_{n+1}|x^n_1)$, using a tokenized LM. In terms of token-level perplexity evaluation, some recent studies \cite{cao2021you, chirkova2023should} have suggested using stochastic tokenization \cite{provilkov2019bpe} at test time to evaluate perplexity scores of LMs ($p(t^i_1)$). However, these evaluations were done on LMs trained with deterministic tokenization which could be suboptimal as demonstrated by our examination of undefined states in Section \ref{bias_analysis}. As such, by utilizing our approach, one can obtain a much more accurate insights on LMs evaluation.


\textbf{Related Algorithms.} Our algorithm is inspired from the literature of universal compression such as prediction by partial matching \cite{cleary1984data} and context-tree weighting \cite{willems1995context}, which have been applied for text prediction but for much simpler settings without any tokenization involved. Recently, \citet{minixhofer2024zero,liu2023task} propose tokenization adaptation methods, which still requires a heuristic optimization that complicates the training pipeline. Some recent studies have proposed method to target the problem of language models encountering difficulties generating text near prompt boundaries \cite{dagan2024getting, guidanceai_github}, which bears some resemblance to our proposed algorithm. These methods, however, are heuristic and only applicable to  certain scenarios. On the other hand, our bias removal algorithm is theoretically correct, versatile for various situations, and enables conversion between token-free and tokenized LMs due to its accurate representation of conditional sampling distributions.

\section{Supporting Theorems on Maximum Prefix Encoding}\label{token_bias_proof}
This section provides supporting theorems for the proof of the main results. We first remind the readers that the set $\mathcal{S}(x^n_1)$ corresponds to the set of all strings that contain $x^n_1$ as a prefix. Similarly, the event set $\mathcal{S}(t^i_1)$ corresponds to the set of all strings whose first $i$ tokens are $t^i_1$. Consider when $t^i_1=\mathrm{encode}(x^{n}_1)$, it should be noted that the two sets $S(t^i_1)$ and $S(x^{n}_1)$ are not guaranteed to be equivalent. That is because the subsequent characters after $x^{n}_1$ can affect the tokenization within the first $n$ character. We illustrate this in more detail in the following example.

\textit{Example.} Consider the Markov chain example in Section \ref{bias_analysis}, where $\mathcal{V}=\{``AA", ``A", ``B"\}$. Then, the string $s_1=``AABAABAB"$, then $s_1\in \mathcal{S}(x_1=``A")$ and $s_1 \in \mathcal{S}(t_1=``AA")$ since the first character of $s_1$ is $``A"$ and the first token of $s_1$ is $``AA"$. On the other hand, $s_1 \notin \mathcal{S}(t_1=\mathrm{encode}(x_1)=``A")$ since its first token is $``AA"$, not $``A"$. 

\begin{figure}[t]
\vspace{-5pt}
\begin{center}
\centerline{\includegraphics[width=0.95\columnwidth]{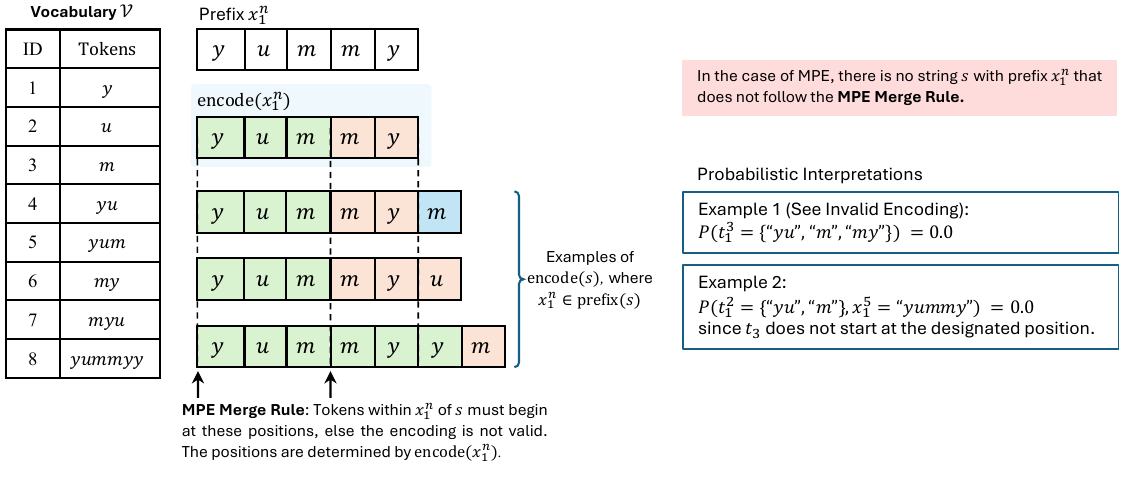}}
\vspace{-5pt}
\captionsetup{font=footnotesize}
\caption{Interpretations of Proposition \ref{tokenbasic}, which shows that for any string $s$ with  prefix $x^n_1$, its token within $x^n_1$ must start at a certain designated positions. For each encoding, same color denotes belonging to the same token. This would later allows us to construct an efficient algorithm to correct the bias. See Corollary \ref{empty_corol} for details on invalid encoding.}
\label{theorems_mpe_illu}
\end{center}
\end{figure}

We introduce the Proposition \ref{tokenbasic} that contains two facts regarding the MPE process, visually presented in Figure \ref{theorems_mpe_illu}.

\begin{proposition} \label{tokenbasic}
Let $s$ be a string with the prefix $x^n_1$ ($x^n_1\in \mathrm{prefix}(s)$). Define the minimal superstring $r$ to be the prefix of $s$ with the fewest tokens that contains $x^n_1$ as a prefix: $r=\mathrm{argmin}_r(k|t^k_1=\mathrm{encode}(r)\wedge x^n_1\in \mathrm{prefix}(r)\wedge r\in\mathrm{prefix}(s))$. Then, we have the followings:
\begin{enumerate}
    \item For $1\leq i < k$, $\mathrm{encode}(s)_i=\mathrm{encode}(x^n_1)_i$. Furthermore, when $r=x^n_1$, we also have $\mathrm{encode}(s)_k=\mathrm{encode}(x^n_1)_k$.
    \item Let $\ell$ be the number of tokens in $\mathrm{encode}(x^n_1)$, then we have $\mathrm{decode}(\mathrm{encode}(x^n_1)^\ell_k) \in \mathrm{prefix}(\mathrm{decode}(\mathrm{encode}(s)_k))$.
\end{enumerate}
\end{proposition}

\begin{proof} 
    \textit{(Result 1.)} Proof by contradiction. Let $s$ be the counter-example with the fewest number of tokens. Assume that for $1\leq i < k$, $\mathrm{encode}(s)_i\neq \mathrm{encode}(x^n_1)_i$. Let $j$ be the smallest of such $i$. 
    
    Consider $\mathrm{encode}(s)_j$ and $\mathrm{encode}(x^n_1)_j$. 
    \begin{itemize}
        \item Case 1: $|\mathrm{decode}(\mathrm{encode}(s)^j_1)| < |x^n_1|$.
        \begin{itemize}
            \item Case 1.a: $|\mathrm{decode}(\mathrm{encode}(s)_j)| < |\mathrm{decode}(\mathrm{encode}(x^n_1)_j)|$. This leads to a contradiction, since $x^n_1$ is a prefix of $s$, therefore a longest prefix matching algorithm would always generate the longer token ($\mathrm{encode}(x^n_1)_j$) over the shorter one ($\mathrm{encode}(s)_j$) when it is available.
            \item Case 1.b: $|\mathrm{decode}(\mathrm{encode}(s)_j)| > |\mathrm{decode}(\mathrm{encode}(x^n_1)_j)|$. This leads to a contradiction, since $\mathrm{concat}(\mathrm{encode}(s)^j_1)$ is a prefix of $x^n_1$ (Case 1 assumption), therefore a longest prefix matching algorithm would always generate the longer token ($\mathrm{encode}(s)_j$)  over the shorter one ($\mathrm{encode}(x^n_1)_j$) when it is available.
            \item Case 1.c: $|\mathrm{decode}(\mathrm{encode}(s)_j)| = |\mathrm{decode}(\mathrm{encode}(x^n_1)_j)|$. This means that the two tokens are the same, contradicting our initial assumption.
        
        \end{itemize}
        \item Case 2: $|\mathrm{decode}(\mathrm{encode}(s)^j_1)| \geq |x^n_1|$. In this case, $r=\mathrm{decode}(\mathrm{encode}(s)^j_1)$ is a superstring of $x^n_1$ implying that $k$ is at most $j$, which contradicts our initial assumption that $1\leq j < k$.
    \end{itemize}

    Finally, in the case $r=x^n_1$, this means $\mathrm{decode}(\mathrm{encode}(s)_k)$ is a suffix of $x^n_1$. Since all the tokens before $k$ within $x^n_1$ has been matched, i.e. $\mathrm{encode}(s)_i=\mathrm{encode}(x^n_1)_i$ for $1\leq i < k$, the last token must also match as the result (else, $|r|\neq |x^n_1|$, leads to contradiction), we have  $\mathrm{encode}(s)_k=\mathrm{encode}(x^n_1)_k$.

    \textit{(Result 2.) } The proof idea is that since $r$ contains $x^n_1$ and any tokens within $r$ and $x^n_1$ has been matched up to $k{-}1$, then what is left in $x^n_1$ must be in the last token in $r$ (which is the $k$th token of $r$). Formally, following Result 1, we have $\mathrm{decode}(\mathrm{encode}(x^n_1)^{k-1}_1) = \mathrm{decode}(\mathrm{encode}(s)^{k-1}_1)$. Since $r$ has $k$ tokens in total and $x^n_1 \in \mathrm{prefix}(r)$, this means that $\mathrm{decode}(\mathrm{encode}(s)_k)$ must cover the rest of $x^n_1$, i.e. $\mathrm{decode}(\mathrm{encode}(x^n_1)^\ell_k)$. As the result, we must have $\mathrm{decode}(\mathrm{encode}(x^n_1)^\ell_k) \in \mathrm{prefix}(\mathrm{decode}(\mathrm{encode}(s)_k))$.
\end{proof}



We remind the reader the definition of invalid encoding below.

\begin{definition}
    (Invalid Encodings) The list of tokens (an encoding) $t^k_1$ is invalid if $\mathrm{encode}(\mathrm{decode}(t^k_1)) {\neq} t^k_1$. Otherwise, it is a valid encoding.
\end{definition}

\begin{corollary}\label{empty_corol}
    $\mathcal{S}(t^k_1)=\emptyset$ if and only if $t^k_1$ is invalid.
\end{corollary}
\begin{proof}
We prove each direction as follow. 
\begin{itemize}
    \item If $\mathcal{S}(t^k_1)=\emptyset$ then $t^k_1$ is invalid: Since $\mathcal{S}(t^k_1)=\emptyset$, we know that there exist no string $s$ such that $\mathrm{encode}(s)^k_1=t^k_1$. As such, for $s=\mathrm{decode}(t^k_1)$, we do not have $\mathrm{encode}(\mathrm{decode}(t^k_1)) = t^k_1$, which proves the result.
    \item If $t^k_1$ is invalid then $\mathcal{S}(t^k_1)=\emptyset$: Let $x^n_1=\mathrm{decode}(t^k_1)$ and $t^i_1 = \mathrm{encode}(x^n_1)$. Let $s_1 \in \mathcal{S}(t^i_1)$ and suppose there exist a string $s_2 \in \mathcal{S}(t^k_1)$. Re-running the MPE procedure on $s_1$ and $s_2$ in parallel, then every time a token is selected within $x^n_1$ in $s_1$, it must also be selected at the same position in $s_2$ as well. Thus, we cannot have $t^i_1 \neq t^k_1$, which proves the result.
\end{itemize}
\end{proof}
\section{Proof of Proposition \ref{undef_prob_main} in the Main Paper}\label{undef_prob_main_proof}
\textbf{Proposition \ref{undef_prob_main}} \textit{   (Token-Induced Zero Probability)
Let $t^i_1$ be a sequence of input tokens. For any invalid encoding $t^i_1$, we have $P_{\mathrm{gt}}(t^i_1){=}0.0$ and the conditional probability $P_{\mathrm{gt}}(t_{i+1}|t^i_1)$ is undefined. In the case $t^i_1$ is valid, then $P_{\mathrm{gt}}(t_{i+1}|t^i_1){=}0.0$ if $t^{i+1}_1$ is invalid.  Furthermore, let $x^{n}_1{=}\mathrm{decode}(t^i_1)$, then for any string $x^{N}_{n+1}$ such that $\mathrm{encode}(\mathrm{concat}(\mathrm{decode}(t^i_1), x^{N}_{n+1})) {\neq} t^i_1$, we have $P_{\mathrm{gt}}(x^{N}_{n+1}|t^i_1){=}0.0.$}

\begin{proof} For the first two statements, we have: 
\begin{itemize}
    \item For  an invalid $t^i_1$ where $t^i_1 \neq \mathrm{encode}(\mathrm{decode}(t^i_1))$, we have $\mathcal{S}(t^i_1)=\emptyset$, as implied by Corollary \ref{empty_corol}. As such, we have $P_{\mathrm{gt}}(t^i_1){=}0.0$ which leads $P_{\mathrm{gt}}(t_{i+1}|t^i_1)$ to be an  undefined conditional probability . 
    \item For a valid $t^i_1$ but invalid $t^{i+1}_1$, we know that  $P_{\mathrm{gt}}(t^{i+1}_1){=}0.0$, which results in $P_{\mathrm{gt}}(t_{i+1}|t^i_1)=0.0$.\end{itemize}
For the last statement, we first note the following:
\begin{enumerate}
    \item Note that $P_{\mathrm{gt}}(x^N_{n+1},t^i_1) =P_{\mathrm{gt}}(x^N_1,t^i_1)$ where $\mathrm{concat}(x^n_1, x^N_{n+1})=x^N_1$.
   \item Consider $P_{\mathrm{gt}}(x^N_1,t^i_1)=P_{\mathrm{gt}}(x^N_1)P_{\mathrm{gt}}(t^i_1|x^N_1)$, we will prove that $P_{\mathrm{gt}}(t^i_1|x^N_1)=0.0$ if $\mathrm{encode}(x^N_1)^i_1\neq t^i_1$.
\end{enumerate}
  The proof idea for this is   shown in Figure \ref{theorems_mpe_illu} (Example 2, Right). Formally:
\begin{itemize}
    \item Let $j$ be the first position such that $\mathrm{encode}(x^N_1)_j\neq t_j$ then we know that $|\mathrm{decode}(\mathrm{encode}(x^N_1)_j))| > |\mathrm{decode}(t_j)|$ (Proposition \ref{tokenbasic} (Result 2)). 
    \item  Following Proposition \ref{tokenbasic} (Result 2), let $s \in \mathcal{S}(x^N_1)$, then we know that $\mathrm{decode}(\mathrm{encode}(x^N_1)_j)$ must be a substring of within another longer token (it cannot be broken down) in $s$. Hence, no string $s$ will have a $j$-th token as $t_j$, so $P_{\mathrm{gt}}(t^i_1|x^N_1)=0.0$. This completes the proof.
\end{itemize}
    


    Finally, we note that $P_\mathrm{gt}(t^i_1) =0.0$ does not implies $\mathrm{encode}(\mathrm{decode}(t^i_1))\in \mathcal{V}$, since it can be due to the original distribution on the character domain. A classic example for this is a Markov model with an absorption state. 
\end{proof}
     
\section{Proof of Proposition \ref{invert_main} and Corollary \ref{backup_equality_main} in the Main Paper}\label{backup_main_proof}

\textbf{Proposition \ref{invert_main}}\textit{ 
    Let $s^*=x^{n}_1$, where $t^i_1=\mathrm{encode}(s^*)=\mathrm{encode}(x^{n}_1)$. Then we have $\mathcal{S}(t^i_1) \subset \mathcal{S}(x^{n}_1)$, i.e. for any string $s$ where $t^i_1=\mathrm{encode}(s)^i_1$,  we have $P(x^{n}_1|t^i_1)=1.0$. In the case $t_i \in \mathcal{V}^*$, then we also have that $\mathcal{S}(t^i_1) = \mathcal{S}(x^{n}_1)$, i.e. any string $s$ where $x^{n}_1\in \mathrm{prefix}(s)$ must have the first $i$ tokens as $t^i_1$ and $P(t^i_1|x^{n}_1)=1.0$. 
 }

\begin{proof} We prove each case as follow.
      
          1) \textit{General Case}: There exists a string $s \in \mathcal{S}(x^{n}_1)$ where $\mathrm{encode}(s)^i_1 \neq t^i_1$, following directly from our 1st order Markov chain example in the main paper, i.e. the string $s=``AA"$ has $``A"$ as prefix but  have the $t_1=``AA" \neq ``A"$. Also, any string $s$ that has the first $i$ tokens as $t^i_1$ must have the first $n$ characters as $x^{n}_1$, hence $\mathcal{S}(t^i_1) \subset \mathcal{S}(x^{n}_1)$ and $P(x^{n}_1|t^i_1)=1.0$.

          2)  $t_i \in \mathcal{V}^*$: The proof idea is that, since $t_i$ cannot be a part of any token in $\mathcal{V}$, it is impossible to merge it by appending additional characters after $t_i$. Formally, similar to Proposition \ref{tokenbasic}:
          \begin{itemize}
              \item For any string $s \in \mathcal{S}(\mathrm{decode}(t^i_1))$, let $\ell$ be the number of tokens in the minimal superstring $r$  of $s$ that contains $x^{n}_1$ as a prefix.
          \item Following Proposition \ref{tokenbasic} (Result 2), we know that $t_i$ must be a substring of $\mathrm{decode}(\mathrm{encode}(s)_\ell)$. 
          \item  Due to $t_i \in \mathcal{V}^*$, then $t_i = \mathrm{encode}(s)_\ell$. We also know from Proposition \ref{tokenbasic} (Result 1) that $\mathrm{encode}(s)_i=t_i$ for $1\leq i < \ell$, this means that $\ell=i$. This gives us $t^i_1=\mathrm{encode}(s)^i_1$ and $P(t^i_1|x^{n}_1)=1.0$.
          \end{itemize}
This completes the proof.           
\end{proof}
\textbf{Remarks. }We briefly note that the condition $t_i\in\mathcal{V}^*$ is the sufficient condition. In general, any token sequence $t^i_1$ that satisfies $\mathcal{S}(t^i_1)=\mathcal{S}(x^n_1)$ will have $P(t^i_1|x^{n}_1)=1.0$. One potential strategy is to find the first index $i=0,1,...k-1$ such that $t^k_{k-i}$ cannot be merged into another token in $\mathcal{V}$.

\textbf{Corollary \ref{backup_equality_main}} \textit{
    Following Proposition \ref{invert_main}, suppose $t_i \in \mathcal{V}^*$ then we have $P(x^N_{n+1}|x^{n}_1{)}  {=} P(x^N_{n{+}1}|t^i_1)$. Similarly, we also have $P(t^j_{i+1}|x^{n}_1{)}  {=} P(t^j_{i+1}|t^i_1)$.}

\begin{proof}
    For the first case, we prove through the following equations:
    \begin{align}
        P(x^N_{n{+}1}|t^i_1) &= P(x^N_{n{+}1}|t^i_1, x^{n}_1) \\
        &= \frac{P(x^N_{n+1}, t^i_1| x^{n}_1)}{P(t^i_1| x^{n}_1)} \\
        &= \frac{P(x^N_{n+1}| x^{n}_1)P(t^i_1| x^{n}_1,  x^N_{n+1})}{P(t^i_1| x^{n}_1)} \\ 
        &= P(x^N_{n+1}| x^{n}_1)
    \end{align}
where the first equality is due to $P(x^{n}_1|t^i_1)=1.0$ and the last equality is due to $P(t^i_1|x^{n}_1)=1.0$ for $t_i \in \mathcal{V}^*$. 

Similarly, for the second case, we have:
\begin{align}
    P(t^j_{i+1}|t^i_1{)} &= P(t^j_{i+1}|x^{n}_1, t^i_1)\\
    &= \frac{P(t^{j}_{i+1}|x^{n}_1)P(t^i_1|x^{n}_1, t^{j}_{i+1})}{P(t^i_1|x^{n}_1)} \\ 
    &= P(t^{j}_{i+1}|x^{n}_1),
\end{align}
which completes the proof.
\end{proof}


\section{Proof for The Bias Removal Method}
\label{BP_proof}

\subsection{Refactoring}
Our goal is to express the quantity $P(x^N_{n+1}|x^n_1)$ using the tokenized LM that outputs the conditional probability $P(t_i|t^{i-1}_1)$. Let $x^{n_k}_1 \in \mathrm{prefix}(x^{n}_1)$ where $t^k_1=\mathrm{encode}(x^{n_k}_1)$ and $t_k \in \mathcal{V}^*$. Following Proposition \ref{invert_main}, any string $s$ with prefix $x^{n_k}_1$ must have the first $k$ tokens as $t^k_1$. We now perform the following factorization:

\begin{align}
    P(x^N_{n+1}|x^n_1) &= P(x^N_{n+1}|x^{n_k}_1, x^n_{n_k + 1})  \\
    &= \frac{P(x^N_{n+1}, x^n_{n_k + 1}|x^{n_k}_1)}{P(x^n_{n_k + 1}|x^{n_k}_1)} \\
    &= \frac{P(x^N_{n_k + 1}|x^{n_k}_1)}{P(x^n_{n_k + 1}|x^{n_k}_1)} \\
    &= \frac{P(x^N_{n_k + 1}|t^{k}_1)}{P(x^n_{n_k + 1}|t^{k}_1)}, \label{fractorial}
\end{align}
where the last inequality is due to Corollary \ref{backup_equality_main}. Finally, we will use the Maximum Prefix Correction (MPE) Algorithm to compute each term in (\ref{fractorial}) individually. Note that the algorithm does not require $t_k \in \mathcal{V}^*$.
Here, we explicitly highlight the importance of having $t_k \in \mathcal{V}^*$, as it bridges between the character and token domain through Equation (\ref{fractorial}).

\subsection{Maximum Prefix Correction Algorithm}
\textbf{Overview. }The MPC algorithm computes $P(x^N_{n_k+1}|t^k_1)$. Note that we do not require $t_k \in \mathcal{V}^*$ in the MPC algorithm. Using marginalization, we have the following:
\begin{align}
    P(x^N_{n_k+1}|t^k_1) &= \sum_{t\in \mathcal{V} }  P(x^N_{n_k+1}, t_{k+1}=t|t^k_1) \\
    &=  \underbrace{\sum_{t{\in}\mathcal{T}_{\mathrm{b_{val}}}}  P(x^N_{n_k+1},  t_{k+1}=t|t^k_1)}_{\mathrm{b_{val}}}  +  \underbrace{\sum_{t{\in}\mathcal{T}_{\mathrm{p_{val}}}}  P(x^N_{n_k+1},  t_{k+1}=t|t^k_1)}_{\mathrm{p_{val}}}\label{bpval_eq}
\end{align}
where:
\begin{itemize}
    \item $\mathcal{T}_{\mathrm{b_{val}}} = \{t \in \mathcal{V} | x^N_{n_k+1} \in \mathrm{prefix}(\mathrm{decode}(t)) \}$ is the set of tokens that have a prefix $x^N_{n_k+1}$. 
    \item $\mathcal{T}_{\mathrm{p_{val}}} = \{ t \in \mathcal{V} | x^N_{n_k+1}\notin \mathrm{prefix}(\mathrm{decode}(t)) \}$ is the ones that do not.
\end{itemize}
 and   $\mathcal{T}_{\mathrm{b_{val}}} \cap \mathcal{T}_{\mathrm{p_{val}}} = \emptyset$. 
 
\textbf{Branch Step. }
Here, $\mathrm{b_{val}}$ is the probability that, given the list of previous tokens $t^k_1$, the next token of the string $s$ has $x^N_{n_k+1}$ as a prefix. To compute this term, we obtain $P(t_{k+1}=t|t^k_1)$ for all $t\in \mathcal{V}$ using one model run, then sum  the probabilities corresponds to all tokens whose prefix is $x^N_{n_k+1}$.
\begin{align}
    \mathrm{b_{val}} = \sum_{t \in \mathcal{T}_{\mathrm{b_{val}}}}  P(t_{k+1}=t|t^k_1), 
\end{align}
\begin{proof}
    To see this, for each summand of $\mathrm{b_{val}}$ in Eq.(\ref{bpval_eq}), we have:
\begin{align}
    P(x^N_{n_k+1}, t_{k+1}=t|t^k_1) &=    P(t_{k+1}=t|t^k_1)\times P(x^N_{n_k+1}|t^k_1, t_{k+1}{=}t)  \\
    &=  P(t_{k+1}=t|t^k_1), 
\end{align}
where $P(x^N_{n_k+1}|t^k_1, t_{k+1}{=}t)=1.0$ is due to  $x^N_{n_k+1} \in \mathrm{prefix}(t)$. This concludes the proof.
\end{proof}

\textbf{Pass Step. } Here, $\mathrm{p_{val}}$ is the probability that, given the list of previous tokens $t^k_1$, the subsequent string $x^N_{n_k+1}$  is \textbf{not} a prefix of the next token. Under the MPE, we compute the value $\mathrm{p_{val}}$ as follow:
  \begin{align}
      \mathrm{p_{val}} = P(t_{k+1}=t|t^k_1)  \times P(x^N_{n_{k+1}+1}|t^k_1, t_{k+1}=t), \label{pass_eq}
  \end{align}
  where $t=\mathrm{encode}(x^N_{n_k+1})_1$ and $x^{n_{k+1}}_{n_k+1} = \mathrm{decode}(t)$. That is, during the passing step, there are two subroutines:
  \begin{enumerate}
      \item Extract the next token $t$ within $x^N_{n_k+1}$ and compute $P(t_{k+1}=t|t^k_1)$. If $x^N_{n_k+1}=\mathrm{decode}(t)$, then returns $0.0$ since this is not allowed according to the condition required in $\mathcal{T}_{\mathrm{p_{val}}}$.  
      \item Recursively compute $P(x^N_{n_{k+1}+1}|t^k_1, t_{k+1}=t)$.
  \end{enumerate}

  \begin{proof}
Following Proposition \ref{undef_prob_main} for invalid encodings, we only need to consider $t$ such that $t^{k+1}_1$ is valid. Under Proposition \ref{tokenbasic} for MPE on $x^N_1$, only first token of $\mathrm{encode}(x^N_{n_k+1})$ is allowed (also see Example 2 in Figure 4(Right)). Finally, applying the chain rule of probability, we obtain Equation \ref{pass_eq}. For the case of non-optimal LM, see  Section \ref{pass_theta} for non-optimal LM.  This completes the proof. \end{proof}

\textbf{Base Case. } We note that the base case of our algorithm corresponds to the situation where $x^N_{n_k+1}=\mathrm{decode}(t)$. In this scenario, we only needs to compute $\mathrm{b_{val}}$ (branching step) while $\mathrm{p_{val}}=0.0$.

\textbf{Complexity Analysis.} The complexity of our algorithm (number of inferences on the language model) scales with the length of the the query string, i.e. $N-n_k$. Note that the complexity of the summation at the Branching step is relatively cheap compared to the runtime of the language model.  

\section{Converting Token-Free Language Model to Tokenized Language Model for MPE.} \label{tkfree2vocab}
We introduce an algorithm to compute $P(t_{k+1}|t^k_1)$ using a token-free language model $P(x^N_{n+1}|x^n_1)$, despite having no access to any tokenized LM. This approach enables theoretical conversion of a token-free model to a tokenized one. The method involves two stages. First, we refactor the conditional probability similar to the technique presented in Section \ref{BP_proof}. Next, we aggregate the probabilities of all possible strings leading to the desired tokenization. It is important to note that a Markov chain is a special type of autoregressive model, meaning this method can be employed to effortlessly calculate Markov chain transition matrices within the tokenized domain.

\subsection{Refactoring}
Consider the probability $P(t_{i+1}|t^i_1)$ that we would like to expressed using $P(x^N_{n+1}|x^n_1)$. Let $t_k$ be the last token within $t^i_1$ such that $t_k \in \mathcal{V}^*$. We now perform the following factorization:

\begin{align}
    P(t_{i+1}|t^i_1) &= \frac{P(t^{i+1}_{k+1}|t^k_1)}{P(t^{i}_{k+1}|t^k_1)} \\
    &= \frac{P(t^{i+1}_{k+1}|x^{n_k}_1)}{P(t^{i}_{k+1}|x^{n_k}_1)},
\end{align}
where $x^{n_k}_1 = \mathrm{decode}(t^k_1)$. The second equality is due to Corollary \ref{backup_equality_main}. Each term can then be computed using the aggregation procedure shown next. 
\subsection{Aggregation.}
In this step, we would like to compute  $P(t^i_{k+1}|x^{n_k}_1)$ where $\mathrm{encode}(x^{n_k}_1) = t^k_1$ and  $t_k \in \mathcal{V}^*$, using the token-free representation $P(x_{n+1}|x^n_1)$. Here, we denote $\mathrm{decode}(t^i_{k+1}) = x^{n_i}_{n_k+1}$ and $M=\max_{t\in\mathcal{V}} |\mathrm{decode}(t)|$ be the length of the longest token in $V$ and $\Omega=\mathcal{A}^M$ is the enumeration of all string of length $M$.

Computing $P(t^i_{k+1}|x^{n_k}_1)$ involves considering all possible strings $s$ with prefix $x^{n_i}_1$ and $t^i_{k+1} = \mathrm{encode}(s)^i_{k+1}$. Although iterating through every possible string is infeasible, we can restrict our search by only examining strings with length $|s|=n_i+M$, as any additional string beyond this point will not impact the tokenization of prefix $x^{n_i}_1 $due to $M$ being the maximum token length. Formally, we will show that one can express $P(t^i_{k+1}|x^{n_k}_1)$ as follows:


\begin{align}
P(t^i_{k+1}|x^{n_k}_1) {=} \sum_{s'\in \mathcal{A}^M} P( x^{n_i+M}_{n_k+1}{=} c_1(s')|x^{n_k}_1)\mathds{1}( t^i_{k+1}{=} \mathrm{encode}(c_2(s'))^i_{k+1}),
\end{align}
where $c_1(s'):=\mathrm{concat}(x^{n_i}_{n_k+1}, s')$ and $c_2(s') := \mathrm{concat}(x^{n_i}_{1}, s')$. The first term can be computed using the given token-free LM, i.e. $P(x^{n_i+M}_{n_k+1}|x^{n_k}_1)$. The second term is an indicator function that checks whether $t^i_{k+1} = \mathrm{encode}(s)^i_{k+1}$ and can be computed deterministically. 
\begin{proof}
    We have:
    \begin{align}
        P(t^i_{k+1}|x^{n_k}_1) &=  P(t^i_{k+1}, x^{n_i}_{n_k+1}|x^{n_k}_1)  \\ 
        &= \sum_{s'\in \mathcal{A}^M} P(t^i_{k+1}, x^{n_i+M}_{n_k+1}{=} c_1(s') |x^{n_k}_1)  \\
        &= \sum_{s'\in \mathcal{A}^M} P( x^{n_i+M}_{n_k+1}{=} c_1(s')|x^{n_k}_1)P( t^i_{k+1}|x^{n_i+M}_1 = c_2(s'))  \\
        &= \sum_{s'\in \mathcal{A}^M} P( x^{n_i+M}_{n_k+1}{=} c_1(s')|x^{n_k}_1)\mathds{1}( t^i_{k+1}{=} \mathrm{encode}(c_2(s'))^i_{k+1})
    \end{align}
    The rest  is to prove the following equality:
    \begin{equation}
        P( t^i_{k+1}|x^{n_i+M}_1 = c_2(s')) = \mathds{1}( t^i_{k+1}{=} \mathrm{encode}(c_2(s'))^i_{k+1})
    \end{equation}    
    We first note that the first $k$ tokens must be $t^k_1=\mathrm{encode}(x^{n_k}_1)$ due to our condition that $t_k \in \mathcal{V}^*$.  Since $M$ is the length of the longest token in $\mathcal{V}$, appending extra characters cannot change the tokenization happened for $x^{n_i}_1$. In other words, any string $s$ with prefix $c_2(s')$ must have the same minimal superstring $r$ containing $x^{n_i}_1$ (see Proposition \ref{tokenbasic}). We then apply this principle to the two cases:
    \begin{itemize}
        \item $t^i_{k+1}{=} \mathrm{encode}(c_2(s'))^i_{k+1}$: In this case, we know that the string must contains the first $i$ tokens as $t^i_{1}$, hence $P( t^i_{k+1}|x^{n_i+M}_1 = c_2(s'))=1.0$
        \item $t^i_{k+1}{\neq} \mathrm{encode}(c_2(s'))^i_{k+1}$: In contrast, this case is equivalent to $P( t^i_{k+1}|x^{n_i+M}_1 = c_2(s'))=0.0$ since we are sure that the string do not contains the tokens $t^i_{k+1}$.
    \end{itemize}
    This concludes the proof.
    \end{proof}
\subsection{The Markov Chain Example.}
We provide a detail computation of the Markov chain example in the main paper. Recall that in the original chain (in the character domain), we have the following:
\begin{align}
 P(x_2=``A"|x_1=``A")&=\alpha \\
 P(x_2=``B"|x_1=``A")&=1-\alpha \\
 P(x_2=``A"|x_1=``B")&=\beta \\
 P(x_2=``B"|x_1=``B")&=1-\beta
\end{align}
We also assume the initial probability $\pi= \{\gamma, 1-\gamma\}$ for $``A"$ and $``B"$ respectively. In the token domain, let first compute $P(t_2=``A"|t_1=``AA")$, where we do not have to do the refactoring step since we know that $t_1\in \mathcal{V}^*$. Following the Aggregation step, we have:
\begin{align}
    P(t_2=``A"|t_1=``AA") &=
    P(x^6_3=``ABA"|x^2_1=``AA") + 
    P(x^6_3=``ABB"|x^2_1=``AA") \\ 
    &= P(x^5_3=``AB"|x^2_1=``AA") \\
    &= \alpha(1-\alpha),
\end{align}
where in the first equality, we do not include the case $x^6_3="AAA"$ and $x^6_3="AAB"$ since $\mathrm{encode}("AAA")_1=``AA"$ and $\mathrm{encode}("AAB")_1=``AA"$, which are not the token $``A"$ that we are interested in. For other tokens and when $t_1=``B"$, the computation follows the same arguments. 

We now consider the case $P(t_2=``B"|t_1=``A")$, we can refactor it as:
\begin{align}
    P(t_2=``B"|t_1=``A")=\frac{P(t_2=``B",t_1=``A")}{P(t_1=``A")}
\end{align}
We first compute $P(t_1=``A")$ using the aggregation step:
\begin{align}
    P(t_1=``A")&=  P(x^3_1=``ABB") + P(x^3_1=``ABA") \\
    &=  P(x^2_1=``AB") \\ 
    &= \gamma (1-\alpha), 
\end{align}
where we do again include the case $x^6_3="AAA"$ and $x^6_3="AAB"$ for the same reason above. For $P(t_2=``A",t_1=``A")$ we have:
\begin{align}
    P(t_2{=}``B",t_1{=}``A")&{=}  P(x^4_1{=}``ABAA") + P(x^4_1{=}``ABAB") + P(x^4_1{=}``ABBA") + P(x^4_1{=}``ABBB") \\
    &{=} P(x^2_1=``AB") \\ 
    &{=} \gamma (1-\alpha) 
\end{align}
which gives us $P(t_2=``B"|t_1=``A")=1.0$. Finally, in this specific case, since order of the Markov chain in the character domain is $1$, we do not need to consider the higher order of the Markov chain in the token domain.





\section{{On Predictive Distribution of Language Models}}\label{lm_justify}

In practice, LMs often do not follow Proposition \ref{undef_prob_main} due to softmax activations. As such, in our MPC algorithm, when $ t\in \mathcal{T}_{\mathrm{p_{val}}}$ and $t \neq \mathrm{encode}(x^N_{n_k+1})_1$, then $P_\theta(x^N_{n_k+1}, t_{k+1}=t|t^k_1)$  may not be $0.0$ (where $\theta$ is the model weights). Eventually, this can potentially increase the complexity of our MPC algorithm during the Passing step. 

In this section, we show that given any tokenized LM, we can force its output probabilities to obey Proposition \ref{undef_prob_main}, without any loss in terms of perplexity score on the token domain. This means that a tokenized LM satisfying Proposition \ref{undef_prob_main} will guarantee the correctness of the Passing step in our MPC algorithm. 

Finally, before going to the method, we remind the readers that Proposition \ref{invert_main} and Corollary \ref{backup_equality_main} are factually correct and hold for all $\theta$. As such, the refactoring step holds regardless. 

\subsection{Truncate-Renormalization Process}
We justify the assumption that our tokenized language model $P_\theta(t_{i+1}|t^i_1)$ follows Proposition \ref{undef_prob_main}. The idea is that we can turn a language model that does not follow Proposition \ref{undef_prob_main} to the one that does while guaranteeing that the new model will always result in a lower token-level perplexity score.  

We first introduce Proposition \ref{renorm_prop}. In this proposition, we are given a target discrete probability distribution $p$ where we know some of the values will not happen, says $\Phi^*$. Assume that we have another distribution $q$ that approximates $p$, then we can produce another distribution $q^*$ that is closer to $p$ in terms of KL divergence by setting corresponding probabilities of $q$ in $\Phi^*$ to $0.0$ and renormalize it  (similar to rejection sampling).
\begin{proposition}\label{renorm_prop}
    Given a discrete distribution $p=\{p_1,p_2,...,p_m\}$ and $q=\{q_1,q_2,...,q_m\}$ with $q_i > 0.0$ for all $i$. Let $\Phi=\{i\in \mathbb{Z}| p_i = 0.0\}$ and $\Phi^* \subseteq \Phi$, we define $q^*=\{q^*_1,q^*_2,...,q^*_m\}$ where $q^*_i = 0.0$ for $ i\in\Phi^*$, and $q^*_j = q_j/(\sum_{l\notin\Phi^*}q_l)$. Then we have: 
    \begin{equation}
        D_{\mathrm{KL}}(p||q^*) \leq D_{\mathrm{KL}}(p||q),
    \end{equation}
    which implies that $q^*$ is closer to $p$ than $q$. We refer to the process of producing $q^*$ as truncate-renormalization (TR). 
\end{proposition}
\begin{proof}
    Let $Z = (\sum_{l\notin\Phi}q_l)$ is the normalizing factor in $q^*$. Note that $Z \leq 1$ and as such $\log(Z) \leq 0$. Then:
    \begin{align}
        D_{\mathrm{KL}}(p||q^*) &= \sum_i p_i \log\left(\frac{p_i}{q^*_i}\right) \\
        &= \sum_{i\notin \Phi^*} p_i \log\left(\frac{p_i}{q^*_i}\right) \quad \text{, use } 0\log0 = 0.0 \\ 
        &= \sum_{i\notin \Phi^*} p_i \log\left(\frac{p_i}{q_i/Z}\right)\\
        &= \left[\sum_{i\notin \Phi^*} p_i \log\left(\frac{p_i}{q_i}\right)\right] + \log(Z) \\
        &\leq \sum_{i\notin \Phi^*} p_i \log\left(\frac{p_i}{q_i}\right) = D_{\mathrm{KL}}(p||q),
    \end{align}
    which completes the proof. 
\end{proof}
Applying to our scenario, for any autoregressive language models $\hat{P}_\theta(t_{i+1}|t^i_1)$ that does not follow Proposition \ref{undef_prob_main} (due to the softmax activations), we can perform the TR process {(since we know which encoding is invalid)} to obtain a new LM  ${P}_\theta(t_{i+1}|t^i_1)$, which is guaranteed to better approximate the ground-truth model ${P_{\mathrm{gt}}}(t_{i+1}|t^i_1)$. Thus, we are guaranteed that the token-level perplexity score of ${P}_\theta(t_{i+1}|t^i_1)$ is always lower than or equal to $\hat{P}_\theta(t_{i+1}|t^i_1)$.  

\subsection{On Passing Step in Maximum Prefix Correction Algorithm.}\label{pass_theta}
Once our tokenized LM follows Proposition \ref{undef_prob_main},  it does not alternate the correctness of the Passing step. In other words, under Proposition \ref{undef_prob_main}, the LM will always output zero probability for invalid encodings $t^k_1$. As a result, the Passing step in the MPC algorithm remains the same in this case.

\section{Algorithms for Byte Pair Encoding}\label{bpe}
\subsection{Overview}
We begin by introducing the Byte-Pair Correction (BPC) Algorithm for bias correction in Byte-Pair Encoding, which is more general than the MPC algorithm and also works for case of MPE. We then follow with a detail analysis to show the correctness of the algorithm.

Here, we introduce the definitions of invalid encodings (for BPE) and cover encodings.

\begin{definition}\label{invalid_enc_bpe}
    (Invalid Encodings) The list of tokens (an encoding) $t^k_1$ is invalid if $\mathrm{encode}(\mathrm{decode}(t^k_1)) {\neq} t^k_1$. Otherwise, it is a valid encoding. We denote a valid $t^k_1$ as $\mathrm{valid}(t^k_1)$.
\end{definition}

\begin{definition}\label{cover_enc_def}
    (Cover Encodings) Given a string $x^n_1$, an encoding $t^k_1$ is said to be covering $x^n_1$ when all the following conditions satisfied:
    \begin{enumerate}
        \item $t^k_1$ is valid.
        \item $x^n_1 \in \mathrm{prefix}(\mathrm{decode}(t^k_1))$.
        \item $x^n_i \in t_k$ for some $1 \leq i \leq n$, i.e. the last token $t_i$ covers a part of the string $x^n_1$.
    \end{enumerate}  We denote $\mathrm{cover}(x^n_1)$ to be the set of all cover encodings of $x^n_1$ and $\Vec{t} \in \mathrm{cover}(x^n_1)$ is an encoding in $\mathrm{cover}(x^n_1)$.
\end{definition}
 Having established these two definitions, we will later show that for BPE (and MPE), the probability $P(x^n_1)$ can be represented using a tokenized LM $P(t_{i+1}|t^i_1)$ as follows: 

\begin{equation}
    P(x^n_1) = \sum_{\Vec{t} \in \mathrm{cover}(x^n_1)} P(\Vec{t}), \label{main_identity}
\end{equation}
and the main goal of the BPC algorithm is to search through all cover encodings of $x^n_1$. We can then apply this algorithm and compute any conditional probability $P(x^N_{n+1}|x^n_1)$ through factorization. Figure \ref{theorems_bpe} (Left) illustrates this with examples cover encodings and invalid/valid encodings.

\subsection{Byte-Pair Correction Algorithm}

\begin{figure}[t]
\vspace{-5pt}
\begin{center}
\centerline{\includegraphics[width=0.95\columnwidth]{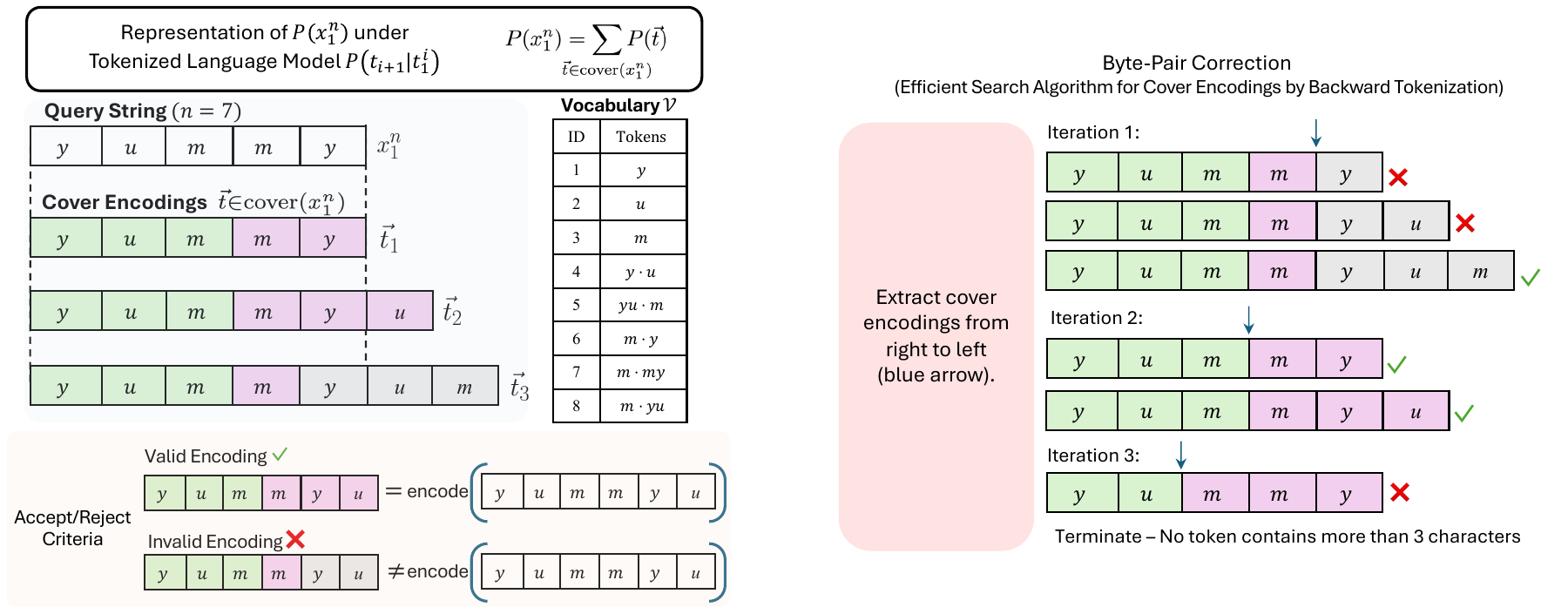}}
\vspace{-5pt}
\captionsetup{font=footnotesize}
\caption{(Left) Representation of $P(x^n_1)$ using tokenized LM. We also show an example of cover encodings and valid/invalid encoding. (Right) Illustration of the Byte-Pair Correction Algorithm for BPE encoding. Green tick and red cross denotes valid and invalid encodings respectively, which can be checked using Definition \ref{invalid_enc_bpe}.  The termination step does not appear in the original algorithm (for simplicity) but can be easily implemented. }
\label{theorems_bpe}
\end{center}
\end{figure}

For MPE, the MPC algorithm computes $P(x^n_1)$ by searching all possible valid encodings that cover $x^n_1$, where the probability of each encoding are computed using the LMs $P(t_{i+1}|t^i_1)$ through greedy search. However, this does not work for the case of BPE. For example, under the BPE encoding rule of Llama 2, the string $\mathrm{``hello"}$ is tokenized as an individual token while the string $\mathrm{``hellow}"$ is tokenized into 2 tokens $\mathrm{``h}$ and $\mathrm{``ellow}"$. Note that naive search for all tokens within $\mathcal{V}$ from left to right that cover $x^n_1$ is computationally expensive.

\begin{algorithm}[h]\footnotesize
\captionsetup{font=footnotesize}
\caption{\footnotesize{Byte-Pair Correction Algorithm. This algorithm computes $P(x^n_1)$ by gradually reducing the search space.}}
\label{algo:BT}
\begin{algorithmic}[1]

\Procedure{compute}{$x^N_1$}       
    \State //\textit{Initialize }$P(x^n_1)$:
    \State $P_{out}=0.0$

    \State //\textit{Probability Aggregation}
    \For{$i =n-1...0$}
    \State //\textit{The last token that partially covers $x^n_1$ begins with $x^n_{i+1}$}
    \State $\mathcal{B}=\{t\in\mathcal{V}|x^n_{i {+} 1} {\in} \mathrm{prefix}( \mathrm{decode} ({t}))\}$

    \State //\textit{Once the last token position is known, the remaining previous tokens can be determined}
    \State $t^{k-1}_1=\mathrm{encode}(x^i_{1})$

    \State //\textit{Similar to the branching step in MPC}
    \State $\mathrm{b_{val}}={\sum\limits_{t {\in }\mathcal{B}}} P(t_{k}=t \big{|} t^{k-1}_1)$
    \State $P_{out}= P_{out} + P(t^{k-1}_1) \times \mathrm{b_{val}}$
    \EndFor\\
    \Return $P_{out}$
\EndProcedure
\end{algorithmic}
\end{algorithm}

The Byte-Pair Correction (BPC) algorithm, shown in  Algorithm \ref{algo:BT} and visualized in Figure \ref{theorems_bpe} (right), which is an efficient algorithm that can search all valid encodings covering $x^n_1$. The idea is that, for each cover encoding $\Vec{t}$, once the starting position of the last token is determined (say $x_{i+1}$), we are guaranteed the prior tokens is unique and must be $\mathrm{encode}(x^i_1)$. Then one will accept the extracted $\Vec{t}$ if it is valid, otherwise reject it. Corollary \ref{empty_corol_bpe} provides justification for this procedure. Here, we assume $P(\Vec{t})=0.0$ for invalid $\Vec{t}$, see Proposition \ref{zero_prob_bpe} and justifications as well as implementation in Appendix \ref{lm_justify}.  

\textbf{Remark. } The BPC algorithm can also be applied for the case of MPE. In fact, it is more general than the original MPC algorithm as it only relies on the property of invalid encodings.

\subsection{Analysis}


\textbf{Notations. } We extend the notation of the vocabulary $\mathcal{V}$ for the case of BPE. Here, $\mathcal{V}$ is an ordered list that determines the merging order in the BPE algorithm. Each $v\in\mathcal{V}$ is a triplet of $(t_\mathrm{left},t_\mathrm{right},t_\mathrm{new})$ which corresponds to the merging tokens (left and right) and the new token. For simplicity, when we write $t \in \mathcal{V}$, it corresponds to the merged token, i.e. $t_\mathrm{new} \in v$. Finally,  the first $|\mathcal{A}|$ entries in $\mathcal{V}$ correspond to the alphabet $\mathcal{A}$, where no merging will happen.

\textbf{Byte-Pair Encoding. }We revise the encoding rule for BPE, shown in Algorithm \ref{algo:BPE}. In practice, pre-tokenization is often used, where tokens are separated by whitespace or special characters. In this case, we can adjust our vocabulary $\mathcal{V}$ by removing tokens with special characters in the middle of the string.


\begin{algorithm}[t]\footnotesize
\captionsetup{font=footnotesize}
\caption{\footnotesize{Byte Pair Encoding Algorithm .}}
\label{algo:BPE}
\begin{algorithmic}[1]

\Procedure{Encode\_BPE}{$x^N_1$, $\mathcal{V}$} 
    \State //\textit{Set initial encodings:}
    \State $\mathrm{c\_tokens} = x^N_1$
    \State //\textit{Iterate over merging order in $\mathcal{V}$, the first $|\mathcal{A}|$ entries correspond the the alphabet (no merge happens):}
    \For{$i =|\mathcal{A}|+1, ...|\mathcal{V}|$}
    \State $\mathrm{c\_tokens} \xleftarrow{} \mathrm{find\_merge}(\mathrm{c\_tokens}, \mathcal{V}[i])$
    \EndFor
\State    \Return $\mathrm{c\_tokens}$ 
\EndProcedure \\

\Procedure{$\mathrm{find\_merge}$}{$\mathrm{c\_tokens}, v$} 
    \State // \textit{Left and right tokens for merging}
    \State $t_{\mathrm{left}}, t_{\mathrm{right}},t_{\mathrm{new}} = v[1], v[2], v[3]$
    \State $\mathrm{old\_tokens} = \mathrm{c\_tokens}$
    \State $\mathrm{new\_tokens} = []$

    \State // \textit{Find and merge tokens from left to right}
    \State $j=1$
    \While{$j < |\mathrm{old\_tokens}|$}
    \If{$ \mathrm{old\_tokens}[i,i+1] = t_{\mathrm{left}}, t_{\mathrm{right}}$} 
    \State $\mathrm{new\_tokens.append}(t_\mathrm{new})$
    \State $j = j + 2$
    \Else
    \State $\mathrm{new\_tokens.append}(\mathrm{old\_tokens}[i])$
    \State $j=j+1$
    \EndIf
    \EndWhile
\State    \Return $\mathrm{new\_tokens}$ 
\EndProcedure
\end{algorithmic}

\end{algorithm}
\textbf{Overview. } We begin our analysis with theoretical results on invalid encodings (Corollary \ref{empty_corol_bpe} and Proposition \ref{zero_prob_bpe}), which characterizes the zero probability events for an optimal tokenized LM. This will allow us to prove the representation of $P(x^n_1)$, i.e. Proposition \ref{marignal_prob_general}, previously shown in Equation (\ref{main_identity}). Finally, we conclude this section with the proof of correctness of the BPC algorithm, using Proposition \ref{marignal_prob_general} and \ref{final_prop}.

We begin with theoretical results on invalid encodings in the case of BPE. 

\begin{corollary}\label{empty_corol_bpe}
    $\mathcal{S}(t^k_1)=\emptyset$ if and only if $t^k_1$ is invalid.
\end{corollary}
\begin{proof}
We prove each direction as follows. 
\begin{itemize}
    \item If $\mathcal{S}(t^k_1)=\emptyset$ then $t^k_1$ is invalid: Since $\mathcal{S}(t^k_1)=\emptyset$, we know that there exist no string $s$ such that $\mathrm{encode}(s)^k_1=t^k_1$. As such, for $s=\mathrm{decode}(t^k_1)$, we do not have $\mathrm{encode}(\mathrm{decode}(t^k_1)) = t^k_1$, which proves the result. 
    \item If $t^k_1$ is invalid then $\mathcal{S}(t^k_1)=\emptyset$: Let $x^n_1=\mathrm{decode}(t^k_1)$, we consider two string $s_1$ and $s_2$ that both have prefix $x^n_1$. Furthermore, we assume the first $i$ tokens of $s_1$ covers exactly $x^n_1$, i.e. $x^n_1=\mathrm{decode}(t^i_1)$ and similarly, the first $j$ tokens of $s_2$ covers exactly $x^n_1$, i.e. $x^n_1=\mathrm{decode}(t^j_1)$. Then:
    
    \begin{enumerate}
        \item Proving invalid $t^k_1$ leads to $\mathcal{S}(t^k_1)=\emptyset$ is equivalently to proving $t^i_1=t^j_1$ for any $s_1,s_2$.
        \item Re-running the BPE algorithm for $s_1$ and $s_2$ in parallel, we know that there will be no merge between any suffix of $x^n_1$ and the rest of strings, i.e. $s_1\backslash x^n_1$ and $s_2\backslash x^n_1$ due to the condition above (See Algorithm \ref{algo:BPE}, line 6).
        \item Furthermore, for $s_1$, any time a merge happens within $x^n_1$ then the same merge must also happen within $x^n_1$ for $s_2$ and vice versa. 
    \end{enumerate}  As the result, we have $t^i_1=t^j_1$ and they must be equal to $\mathrm{encode}(x^n_1)$.
\end{itemize}
\end{proof}


\begin{proposition}\label{zero_prob_bpe}(Token-Induced Zero Probability-BPE)
    Let $t^i_1$ be a sequence of input tokens. For any invalid encoding $t^i_1$, we have $P_{\mathrm{gt}}(t^i_1){=}0.0$ and the conditional probability $P_{\mathrm{gt}}(t_{i+1}|t^i_1)$ is undefined. In the case $t^i_1$ is valid, then $P_{\mathrm{gt}}(t_{i+1}|t^i_1){=}0.0$ if $t^{i+1}_1$ is invalid.  
\end{proposition}

\begin{proof} The proof is the same as the MPE version (Proposition \ref{undef_prob_main}). 
\end{proof}




\textbf{Correctness of BPC Algorithm. }We show in Proposition \ref{marignal_prob_general} that computing the string probability $P(x^n_1)$ is equivalent to marginalizing the probability of all covering tokens of $x^n_1$. As such, the main task of computing $P(x^n_1)$ is to iterate all the valid encodings that cover $x^n_1$.

\begin{figure}[t]

\begin{minipage}[b]{0.45\textwidth}
\centering
\includegraphics[width=1\textwidth]{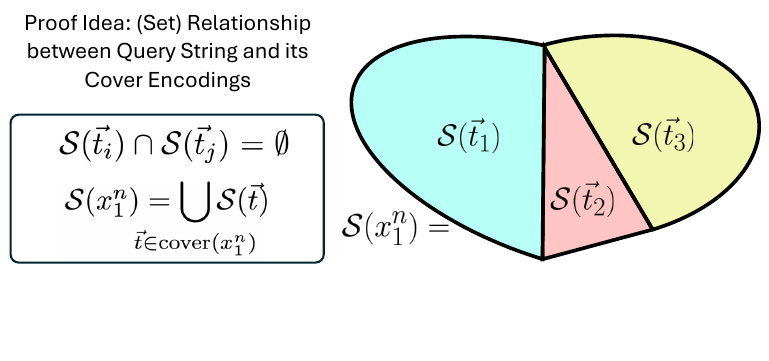}
\vspace{-30pt}
\captionsetup{font=footnotesize}
\caption{Illustration of Proposition \ref{marignal_prob_general} when $|\mathrm{cover}(x^n_1)|{=}3$.}
\label{Proof BPE Visualization}
\end{minipage}
\hfill
\begin{minipage}[b]{.45\textwidth}
\centering
\includegraphics[width=1\textwidth]{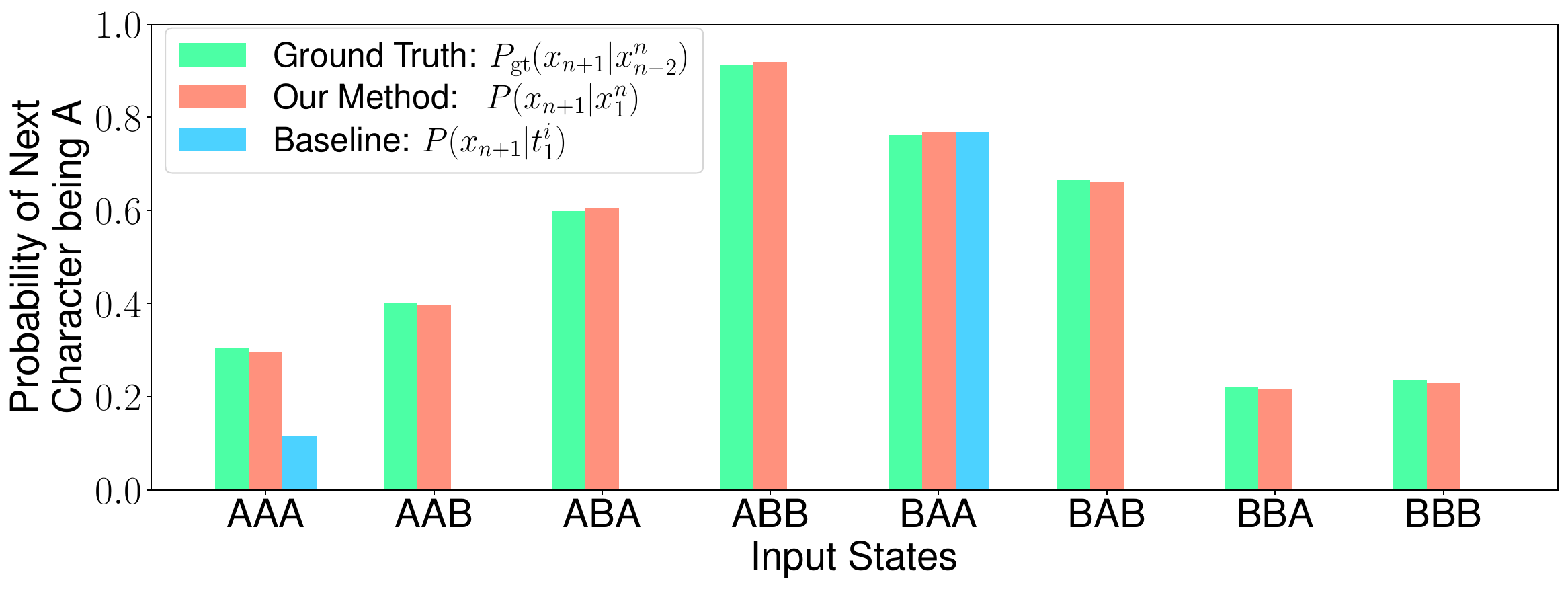}
\vspace{-20pt}
\captionsetup{font=footnotesize}
\caption{Our method accurately estimates the transition probability of a 3rd order Markov chain while the baseline method fails to.}\label{results_bpe}
\end{minipage}
\vspace{15pt}

\end{figure}

\begin{proposition}\label{marignal_prob_general} (Prefix Probability Representation)
    Given a prefix $x^n_1$, we have the followings:
    \begin{enumerate}
        \item For any distinct $\Vec{t}_i, \Vec{t}_j \in \mathrm{cover}(x^n_1)$, then $\mathcal{S}(\Vec{t}_i) \cap \mathcal{S}(\Vec{t}_j){=}\emptyset$.
        \item $\mathcal{S}(x^n_1)=\smashoperator{\bigcup_{\Vec{t} \in \mathrm{cover}(x^n_1)}} \mathcal{S}(\Vec{t})$.
    \end{enumerate}
       As a result,  $P(x^n_1)$ can be expressed as the marginal probability of all covering tokens of $x^n_1$
    \begin{equation}
        P(x^n_1) = \sum_{\Vec{t} \in \mathrm{cover}(x^n_1)} P(\Vec{t})
    \end{equation}
\end{proposition}
\begin{proof} 


    We prove each point as follows:
    \begin{enumerate}
        \item Proof by contradiction, suppose that $\mathcal{S}(\Vec{t}_i) \cap \mathcal{S}(\Vec{t}_j)\neq \emptyset$, then there exists a string $s$ that has two different cover encodings $\Vec{t}_1$ and $\Vec{t}_2$. This is impossible since each string $s$ has only one unique encoding.
        \item This follows the definition of cover encodings.
    \end{enumerate}
    Since each $\mathcal{S}(\Vec{t})$ is pair-wise disjoint, we arrive at the final equation. We illustrate this Proposition in Figure \ref{Proof BPE Visualization}.
\end{proof}

Finally, we prove that BPC extracts all the cover encodings of $x^n_1$. Proposition \ref{final_prop} shows the correctness of line 9 in the algorithm, where suppose that the last token starts from $x_{i+1}$, then $\mathrm{encode}(x^i_1)$ must be the encoding before that last token. Since each cover encoding $\Vec{t}$ must have a last token cover a suffix within $x^n_1$, iterating over all positions from $1$ to $n$ guarantees that we extract all possible encodings.

\begin{proposition}\label{final_prop}
    Let $\Vec{t}  \in \mathrm{cover}(x^N_1)$ and $k=|\Vec{t}|$ is the number of tokens in $\Vec{t}$. Suppose that $x^N_{j+1}$ is a prefix of the $t_k$ for some $0\leq j \leq N$, i.e. $x^N_j \in \mathrm{prefix}(\Vec{t}_{k})$, then $t^{k-1}_1 = \mathrm{encode}(x^{j}_1)$. 
\end{proposition}

\begin{proof}
    Since $\Vec{t} = t^k_1 \in \mathrm{cover}(x^N_1)$, then $t^{k-1}_1$ must be a valid encoding. As a result, we must have $t^{k-1}_1=\mathrm{encode}(x^j_1)$.
\end{proof}

\textbf{Refactoring. }Unlike the case for MPE, identifying the case when $P(x^N_{n+1}|t^i_1)=P(x^N_{n+1}|x^n_1)$ is nontrivial in general for BPE. Nevertheless, the refactoring step aims to reduce the computational complexity, and in general, we can still compute $P(x^N_{n+1}|x^n_1)$ by refactoring. 
\begin{align}
    P(x^N_{n+1}|x^n_1)=\frac{P(x^N_1)}{P(x^n_1)},
\end{align}
and we use the BPC algorithm to compute $P(x^N_1)$ and $P(x^n_1)$ respectively. Note that this is equivalent to assuming $t_1$ is a ${<}\mathrm{start}{>}$ token within $\mathcal{V}^*$ (and consider $P(x^N_2|t_1), P(x^n_2|t_1)$ instead of $P(x^N_1),P(x^n_1)$). When pretokenization is used, e.g. split by white spaces, we can identify when $P(x^N_{n+1}|t^i_1)=P(x^N_{n+1}|x^n_1)$ by using the pretokenization pattern. 


\subsection{Experiments}
The experiment setup for BPE is the same as the one in Section \ref{mpe_exp}, except we use the vocabulary $\mathcal{V} = \{``A", ``B", ``B {\cdot} A", ``BA{\cdot}A", ``B{\cdot}BAA", ``A{\cdot}A", ``BA{\cdot}BA", ``B{\cdot}B"\}$, where the order within $\mathcal{V}$ is the merging order for the BPE encoding process and the ``${\cdot}$" separates the merging tokens. The result is shown in Figure \ref{results_bpe}, where our method can accurately recover the ground truth probability $P(x_{n+1}|x^n_1)$ while the baseline fails to. Notice that for the state $``BAA"$, the baseline approach can output the correct probability, which is because the merging for the token $``BAA"$ happens before any mergs where $``A"$ is the left token happens.  This experiment also shows the existence of bias within the BPE process and our method can recover the exact ground truth probability.

\end{document}